\documentclass[10pt,twocolumn,letterpaper]{article}

\usepackage{iccv}
\usepackage{times}
\usepackage{epsfig}
\usepackage{graphicx}
\usepackage{amsmath}
\usepackage{amssymb}
\usepackage{enumitem}
\usepackage{listings}
\usepackage{ragged2e}
\usepackage{xcolor}
\usepackage{placeins}
\definecolor{deepblue}{rgb}{0,0,0.5}
\definecolor{deepred}{rgb}{0.6,0,0}
\definecolor{deepgreen}{rgb}{0,0.5,0}
\newcommand\pythonstyle{\lstset{
language=Python,
basicstyle=\small\ttfamily,
morekeywords={self},              %
keywordstyle=\small\ttfamily\bfseries\color{deepblue},
emph={lm,concat,captioner,argmax,append,zeros_like,logsumexp},          %
emphstyle=\small\ttfamily\bfseries\color{deepred},    %
stringstyle=\color{deepgreen},
frame=tb,                         %
showstringspaces=false
}}

\lstnewenvironment{python}[1][]
{
\pythonstyle
\lstset{#1}
}
{}

\usepackage{amsmath,amsfonts,bm}

\def\eqref#1{equation~\ref{#1}}

\def\1{\bm{1}}

\def\rx{{\textnormal{x}}}
\def\ry{{\textnormal{y}}}

\def\vzero{{\bm{0}}}

\def\vtheta{{\bm{\theta}}}

\def\vc{{\bm{c}}}

\def\vv{{\bm{v}}}

\DeclareMathAlphabet{\mathsfit}{\encodingdefault}{\sfdefault}{m}{sl}
\SetMathAlphabet{\mathsfit}{bold}{\encodingdefault}{\sfdefault}{bx}{n}

\newcommand{\softmax}{\mathrm{softmax}}

\iccvfinalcopy

\usepackage{xcolor}
\usepackage{chngcntr}

\newcommand{\citep}[1]{\cite{#1}}
\newcommand{\citet}[1]{\cite{#1}}
\DeclareRobustCommand{\tablefont}{%
        \fontencoding{\encodingdefault}%
        \fontseries{m}%
        \fontshape{n}%
        \fontfamily{phv}%
        \fontsize{4.5}{5.75}%
        \selectfont}
\DeclareTextFontCommand{\texttable}{\tablefont}%

\usepackage[pagebackref=true,breaklinks=true,letterpaper=true,colorlinks,bookmarks=false]{hyperref}

\def\iccvPaperID{11082} %

\ificcvfinal\fi

\begin{document}
\setlength{\abovedisplayskip}{4pt}
\setlength{\belowdisplayskip}{4pt}
\title{Guiding image captioning models toward more specific captions}

\author{
\makebox[\textwidth]{
\hspace*{\fill}
Simon Kornblith\textsuperscript{1}
\hfill
Lala Li\textsuperscript{1}
\hfill
Zirui Wang\textsuperscript{2}\thanks{Work performed while at Google.}
\hfill
Thao Nguyen\textsuperscript{3}\thanks{Work performed as a student researcher at Google.}
\hspace*{\fill}
}
\\
\textsuperscript{1}Google DeepMind \qquad \textsuperscript{2}Apple AI/ML \qquad \textsuperscript{3}University of Washington
}

\newcommand{\fix}{\marginpar{FIX}}
\newcommand{\new}{\marginpar{NEW}}

\maketitle
\ificcvfinal\thispagestyle{empty}\fi

\begin{abstract}
Image captioning is conventionally formulated as the task of generating captions for images that match the distribution of reference image-caption pairs. However, reference captions in standard captioning datasets are short and may not uniquely identify the images they describe. These problems are further exacerbated when models are trained directly on image-alt text pairs collected from the internet. In this work, we show that it is possible to generate more specific captions with minimal changes to the training process. We implement classifier-free guidance~\citep{ho2021classifierfree} for an autoregressive captioning model by fine-tuning it to estimate both conditional and unconditional distributions over captions. The guidance scale applied at decoding controls a trade-off between maximizing $p(\mathrm{caption}|\mathrm{image})$ and $p(\mathrm{image}|\mathrm{caption})$. Compared to standard greedy decoding, decoding with a guidance scale of 2 substantially improves reference-free metrics such as CLIPScore (0.808 vs. 0.775) and caption$\to$image retrieval performance in the CLIP embedding space (recall@1 44.6\% vs. 26.5\%), but worsens standard reference-based captioning metrics (e.g., CIDEr 78.6 vs 126.1). We further explore the use of language models to guide the decoding process, obtaining small improvements over the Pareto frontier of reference-free vs. reference-based captioning metrics that arises from classifier-free guidance, and substantially improving the quality of captions generated from a model trained only on minimally curated web data. 
\end{abstract}

\begin{figure}
    \centering
    \setlength{\tabcolsep}{2pt}
    \begin{tabular}[t]{lr}
    \begin{minipage}{0.48\linewidth}
    \includegraphics[width=1.0\linewidth]{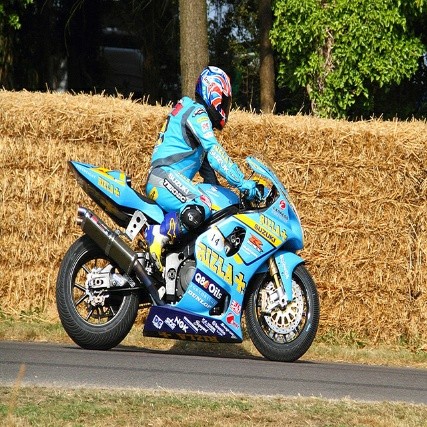}
    \tablefont
    \begin{tabular}[t]{rp{0.82\linewidth}}
$\gamma$=1.0& a man riding a blue motorcycle on a dirt road\\
$\gamma$=1.5& a man riding a blue motorcycle on a straw bale\\
$\gamma$=2.0& rider on blue suzuki motorcycle near straw bales\\
$\gamma$=3.0& racer on blue suzuki motorcycle leaning up against straw bales\\
GT& A person riding a baby blue motorcycle near haystacks
    \end{tabular}
    \end{minipage}
    &
    \begin{minipage}{0.48\linewidth}
    \includegraphics[width=1.0\linewidth]{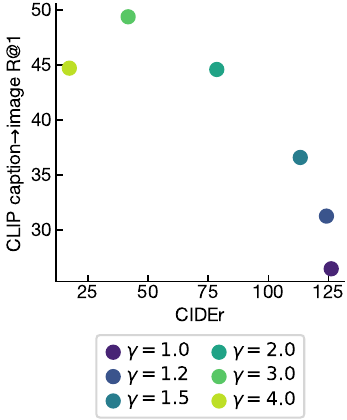}
    \end{minipage}
    \end{tabular}
    \vspace{0.5em}
    \caption{Using classifier-free guidance ($\gamma>1$) results in more specific captions that are farther from the reference distribution. Left: Example of captions generated at different guidance scales for a single image. Right: Caption$\to$image recall@1 with CLIP ViT-B/32 vs. CIDEr score, for captions generated with different guidance scales $\gamma$ on MS-COCO. Higher scales improve retrieval accuracy at the expense of CIDEr.}
    \label{fig:my_label}
    \vspace{-1em}
\end{figure}

\section{Introduction}

Image captioning is both a difficult task for computer vision systems to perform and a difficult task to evaluate. Although automated captioning metrics rank the best captioning systems higher than humans, human raters still show a strong preference for human-generated captions~\citep{kasai2021transparent}, suggesting shortcomings in both captioning models and metrics. One shortcoming relates to the lack of specificity in generated captions. Conventional maximum likelihood-based image captioning models attempt to generate captions such that the $p(\mathrm{caption}|\mathrm{image})$ is high. However, captions from the ground truth distribution are often non-specific, e.g., human annotators will usually describe a German Shepard only as a dog. Moreover, previous work has emphasized ``reference-based'' captioning metrics that measure the match between generated captions and human-provided ground truth captions~\citep{papineni2002bleu,lin2004rouge,vedantam2015cider}. These metrics intrinsically penalize captions that are more specific than ground truth. %

In this work, we explore strategies to guide image captioning models to produce more specific captions by modifying the decoding distribution, and explore the trade-offs in captioning metrics that result. We first investigate the application of classifier-free guidance (CFG)~\citep{ho2021classifierfree} to image captioning with autoregressive models. Classifier-free guidance increases $p(\mathrm{image}|\mathrm{caption})$ at the expense of $p(\mathrm{caption}|\mathrm{image})$. Although CFG hurts reference-based image captioning metrics such as BLEU~\citep{papineni2002bleu}, ROUGE~\citep{lin2004rouge}, and CIDEr~\citep{vedantam2015cider},
it improves ``reference-free'' metrics that measure captions' specificity via the similarity between the image and the generated caption in the embedding space of image-text models~\citep{hessel-etal-2021-clipscore} or caption$\to$image retrieval performance. Qualitatively, we find that captions generated with CFG are more specific than both the ground truth captions and captions generated without CFG, but they are less grammatical, particularly at high CFG scales.

Beyond classifier-free guidance, we experiment with guiding image captioning models using the probability distribution obtained from a few shot-prompted language model (LM). We find that using a language model to guide a captioning model trained on MS-COCO~\cite{lin2014microsoft} with descriptive manually written captions can allow it to achieve slightly better trade-offs between reference-free vs. reference-based captioning metrics than those observed with CFG. LM guidance also substantially improves the captions produced by a model trained exclusively on minimally curated web data. %
Although this model achieves a CIDEr score of only 21.8 without guidance, this CIDEr score improves to 57.4 when guided by a language model prompted with 20 captions from the MS-COCO training set.

In summary, our contributions are as follows:
\begin{itemize}[leftmargin=1em,itemsep=0.3em,parsep=0em,topsep=0.3em]
    \item We propose two strategies to guide image captioning models to produce more specific captions: classifier-free guidance and language model guidance.
    \item We demonstrate that classifier-free guidance yields captions that are closer to the corresponding image in the embedding space of image-text models, but are farther from human-provided reference captions.
    \item We show that language model guidance can alter caption styles, substantially improving captions produced by a model trained only on minimal curated web data and marginally improving the trade-off between captioning metrics observed with classifier-free guidance.
\end{itemize}

\vspace{-0.5em}
\section{Related work}
\textbf{Measuring specificity of captions.} Early work using neural networks for image captioning found that models have a propensity to regurgitate captions from their training data, and as a result, the generated captions are not descriptive enough to uniquely identify images~\citep{vinyals2015show,devlin2015language}. 
To address this shortcoming, Lindh et al. \citet{lindh2018generating} proposed to use caption$\to$image recall with an image retrieval model to examine whether images can be retrieved from generated captions, and further attempt to differentiate through this retrieval process to train a captioning model. %
Their approach marginally improves retrieval accuracy, but worsens reference-based captioning metrics. More recent work has adopted approaches to evaluate  the specificity of captions based on the CLIP image-text model~\citep{radford2021learning}. Hessel et al. \citet{hessel-etal-2021-clipscore} propose CLIPScore, an image captioning metric based on the cosine similarity between CLIP embeddings of the image and the generated caption. Kasai et al. \citet{kasai2021transparent} report that CLIPScore-based metrics align better with human judgments compared to reference-based captioning metrics.

\textbf{Improving specificity of captions.} Recent work has attempted to directly optimize CLIP-based losses that measure the similarity of captions with corresponding images in the CLIP embedding space, either on their own or jointly with CIDEr scores. Work that trains captioning models has generally approached this problem using reinforcement learning, and finds that adding these losses worsens standard reference-based captioning metrics but improves similarity and retrieval in the CLIP embedding space~\citep{huang2021unifying,cho2022fine,zhang2022distincive}, similar to our observations regarding CFG. Wen et al.~\citep{wen2023hard} attempt to generate prompts for text-to-image generative models that correspond to specific images without a captioning model, by directly optimizing the similarity between the text and image in the CLIP embedding space using a gradient-based discrete optimization procedure, but the resulting text is not grammatical.

Other work has attempted to generate more descriptive captions through different means. Dense captioning~\cite{yang2017dense} aims to detect and caption all objects in an image, but concatenating all of these captions leads to long and unnatural captions, whereas CFG produces single-sentence captions. The Localized Narratives dataset~\cite{pont2020connecting} contains visually grounded captions for MS-COCO images collected through voice annotation. These captions are substantially more descriptive than the captions in the MS-COCO dataset and can be used for model training. Concurrent with our work, IC$^3$~\citet{chan2023ic} proposes to generate multiple captions with an off-the-shelf captioning model and combine them using a language model. The resulting captions are longer, but achieve greater caption$\to$image recall. %

\textbf{Captioning from uncurated data.} In Section~\ref{sec:lm_guidance}, we explore the use of LM guidance for captioning with access to uncurated image-text data from the web and a small number of captions but not images from the target distribution. This setting, which does not rely on aligned images and captions from the target distribution, is often referred to as ``zero-shot'' captioning, and previous work has pursued a number of alternative approaches. Flamingo~\cite{alayrac2022flamingo} and CM3~\cite{aghajanyan2022cm3} perform zero-shot captioning by pretraining on interleaved image/text data. MAGIC~\cite{su2022language} and ZeroCap~\cite{tewel2022zerocap} generate captions using a combination of guidance from CLIP and a large language model. Other recent work adapts CLIP to perform captioning by training a text decoder using only captions, with no corresponding images~\cite{nukrai2022text,lidecap}.

\textbf{Classifier-free guidance.} CFG is widely used in diffusion-based and autoregressive text-to-image models~\citep{nichol2021glide,ramesh2022hierarchical,sahariaphotorealistic,rombach2022high,gafni2022make,yu2022scaling}. Because of the popularity of the combination of CFG and diffusion, previous work that has performed image captioning with diffusion models has also examined the use of CFG. 
This work finds either no benefit to using CFG \citep{xu2022clip} or a small and inconsistent benefit that appears to vary with minor changes in training settings~\citep{zhu2022exploring}. However, these studies do not seek to generate more specific captions, and thus measure only reference-based captioning metrics, which we likewise find do not benefit from CFG. %
Concurrently with out work, \citet{sanchez2023stay} propose to use classifier-free guidance to improve prompt following in large language models. %

\section{Methods}

\subsection{Classifier-free guidance for image captioning}
Let $\rx$ be an image caption and $\ry$ be the corresponding image. A standard captioning model aims to model the likelihood $p(\rx|\ry)$, factorized autoregressively in terms of the probability of each token given previous tokens
\begin{align}
    p(\rx|\ry) = p(\rx_n|\rx_{n-1},\dots\rx_1,\ry) \dots p(\rx_1|\ry).
\end{align}
The network is trained so that its output distribution $q_\vtheta(\rx_n | \rx_{n-1},\dots\rx_1,\ry) \mathrel{\overset{\scriptscriptstyle \mathrm{def}}{=}}
 \softmax(f_\vtheta(\rx_{n-1},\dots\rx_1,\ry))$ approximates  $p(\rx_n|\rx_{n-1},\dots\rx_1,\ry)$. At inference time, one typically uses beam search or greedy decoding to produce a caption that has a particularly high probability. In this work, we use greedy decoding because it is the more common choice and it is also simpler to implement.

Classifier-free guidance (CFG)~\citep{ho2021classifierfree} aims to generate outputs that maximize or otherwise achieve high values of
\begin{align}
    l_{\vtheta,\gamma}(\rx,\ry) \mathrel{\overset{\scriptscriptstyle \mathrm{def}}{=}} p(\rx)\left(\frac{p(\rx|\ry)}{p(\rx)}\right)^\gamma \propto p(\rx)p(\ry|\rx)^\gamma
    \label{eq:cfg}
\end{align}
where proportionality holds because $p(\rx|\ry)/p(\rx) = p(\ry|\rx)/p(\ry)$ and $p(\ry)$ is fixed. The parameter $\gamma$ is called the guidance scale and controls the trade-off between maximization of $p(\rx|\ry)$ and $p(\ry|\rx)$. When $\gamma = 1$, $l_{\vtheta,\gamma}(\rx,\ry) = p(\rx|\ry)$ and guidance has no effect. Setting $\gamma > 1$ inflates the probability of the image given the caption $p(\ry|\rx)$ relative to the unconditional probability of the caption $p(\rx)$.

Ho and Salimans \citet{ho2021classifierfree} originally proposed CFG in the context of diffusion models, which estimate the score functions $\nabla \log p(\rx|\ry)$ and $\nabla \log p(\rx)$. Although $l_{\vtheta,\gamma}(\rx,\ry)$ factorizes autoregressively, it is not a normalized probability distribution, so it is not entirely clear how one should sample tokens when performing autoregressive generation. Crowson \citet{crowson2022classifierfree} suggested to sample from
\begin{flalign} 
    &\tilde{q}_{\vtheta,\gamma}(\rx_{n}|\rx_{n-1},\dots,\rx_{1},\ry) \mathrel{\overset{\scriptscriptstyle \mathrm{def}}{=}} \softmax( f_\vtheta(\rx_{n-1},\dots,\rx_{1},\vzero)\nonumber\\
    &\hspace{1.0em}+ \gamma(f_\vtheta(\rx_{n-1},\dots,\rx_{1},\ry) - f_\vtheta(\rx_{n-1},\dots,\rx_{1},\vzero))),
\label{eq:something}
\end{flalign}
where $f_\vtheta(\rx_{n-1},\dots,\rx_{1},\vzero)$ are logits generated by the model without conditioning, usually by passing zeros in place of the conditioning information. This formulation has been successfully applied in autoregressive image models~\citep{gafni2022make,yu2022scaling}. In our experiments, we adopt this formulation as well, but since we decode greedily, i.e., at each step we take the token that maximizes $\tilde{q}_{\vtheta,\gamma}(\rx_{n}|\rx_{n-1},\dots,\rx_{1},\ry)$ and thus $l_{\vtheta,\gamma}(\rx,\ry)$, any form of normalization of the per-step sampling distribution would produce the same captions. We provide pseudocode in Appendix~\ref{app:pseudocode_cfg}.

\subsection{Language model guidance}
Inspired by classifier-free guidance, we consider \textit{language model (LM) guidance}, which attempts to maximize
\begin{align}
    l'_{\vtheta,\gamma}(\rx,\ry) \mathrel{\overset{\scriptscriptstyle \mathrm{def}}{=}} q(\rx)\left(\frac{p(\rx|\ry)^\alpha}{p(\rx)^\beta}\right),
    \label{eq:lmg}
\end{align}
where $p(x)$ and $p(x|y)$ are obtained from a captioning model as in CFG but $q(\rx)$ is obtained from a language model that was trained independently (but with the same vocabulary) on a large text corpus. The quantity $p(\rx|\ry)/p(\rx) = p(\rx,\ry)/(p(\rx)p(\ry))$ measures the strength of the association between a caption and an image; its logarithm is the pointwise mutual information (PMI). LM guidance relies on the assumption that, even for large shifts in the prior distribution of captions $p(\rx)$, the shift in PMI will be small. Empirically, we  obtain better results by allowing different exponents for the numerator and denominator, with $\alpha > \beta$. This decoupling resembles PMI$^k$~\citep{daille1994approche}, which reduces the bias of PMI toward rare associations. We provide a more detailed derivation in Appendix~\ref{app:derivation_lmg}.

We investigate two applications of LM guidance. First, we combine a captioning model fine-tuned on MS-COCO with a LM prompted with manually written descriptive captions to alter the style of the captions the model produces. The manually written prompts are shown in Appendix~\ref{app:captions}. Second, we combine a captioning model trained only on low-quality web data with a LM prompted with varying numbers of examples from the MS-COCO training set to evaluate the ability of LM guidance to elicit higher-quality captions without high-quality paired data. We randomly select a different set of captions for each minibatch of four test examples. In both cases, we separate the captions with two newlines. Because this format leads the LM to place probability mass on the newline token to end the caption, we transfer the probability mass from the newline token to the EOS token. See Appendix~\ref{app:pseudocode_lmg} for pseudocode.

\subsection{Models and training}
\label{sec:models_and_training}
Our captioning model is a ``bottleneck'' variant of CoCa-Base~\citep{yu2022coca}, which combines a contrastive loss with a captioning loss to simultaneously learn aligned image and text embeddings as well as a captioner. The architecture consists of an image encoder, a unimodal text decoder, and a multimodal text decoder, each of which are Transformers with 12 layers, 768 hidden dimensions, an MLP of size 3072, and 12 self-attention heads, matching  BERT\textsubscript{BASE}~\cite{devlin-etal-2019-bert} and GPT-1~\cite{radford2018improving}. The image encoder is a ViT-B/18 that processes $288\times 288$ input and produces an embedding such that images are embedded close to their corresponding text.

CoCa's multimodal text decoder processes the representations of the image encoder to produce a caption. Whereas \citet{yu2022coca} conditions the multimodal text decoder using cross-attention to pooled representations, our bottleneck variant uses only the contrastive image embedding. Appendix~\ref{app:architecture} shows a diagram of the resulting architecture. We adopt this bottleneck variant because of its simplicity and the conceptual appeal: When CFG is used, the captioner's role is to invert the image embedding, providing a caption that, when embedded by the text encoder, lies close to it. However, as we show in Appendix~\ref{app:coca_no_bottleneck}, this choice of the bottleneck model is not critical, and CFG is equally effective with the standard CoCa architecture with attention pooling.

For CFG experiments, we pretrain our model on an image-text dataset comprising images from the JFT-5B dataset~\citep{sun2017revisiting,zhai2022scaling} paired with their corresponding label names substituted into a randomly selected prompt from the list provided by Radford et al.~\citep{radford2021learning}, web images paired with noisy alt text from the ALIGN dataset~\citep{jia2021scaling}, and a small amount of data from other sources.
We follow the same recipe as in~\citet{yu2022coca}, and do not mask conditioning information during pretraining.\footnote{We find that passing an all-zero image embedding to the pretrained model yields samples that resemble the unconditional distribution, suggesting that it implicitly learns to model the unconditional distribution.} We then fine-tune on the combined MS-COCO train and Karpathy validation splits~\cite{karpathy2015deep} using Adam with batch size 128. We linearly warm up to a learning rate of $1 \times 10^{-5}$ over the first $1,000$ steps and linearly decay to zero over the rest of training. We vary $\gamma \in \{1.0, 1.2, 1.5, 2.0, 3.0, 4.0\}$, conditioning masking proportion in $\{0.0, 0.25, 0.5, 0.75\}$, and numbers of steps in $\{5{,}000, 10{,}000, 20{,}000, 50{,}000\}$.  We report results from the model trained for $20{,}000$ steps with  masking proportion $0.5$, which achieves near-optimal results, in Tables \ref{tab:results} and \ref{tab:coca_model_metrics}, and sample example captions from it. To ensure that results generalize across datasets, we also experiment with a model fine-tuned on Conceptual Captions~\cite{sharma2018conceptual} for $100{,}000$ steps with masking proportion $0.5$.

For LM guidance experiments, we pretrain on the JFT-5B and ALIGN datasets, again following the recipe of \citet{yu2022coca}. For zero-shot captioning experiments, we fine-tune this model on the same datasets for an additional 50,000 steps with conditioning masking proportion of 0.5 to improve our ability to sample unconditionally. For LM guidance on MS-COCO, we first fine-tune on ALIGN, JFT-5B images backcaptioned by an MS-COCO fine-tuned CoCa-2B model, and a small amount of internal data before fine-tuning on MS-COCO. Our language model is a variant of Primer~\citet{so2021primer} with 2 billion parameters, trained on a similar dataset to that used to train PaLM~\citep{chowdhery2022palm}.

\subsection{Evaluation}

\begin{figure*}[t]
\begin{center}
\includegraphics[width=0.82\linewidth]{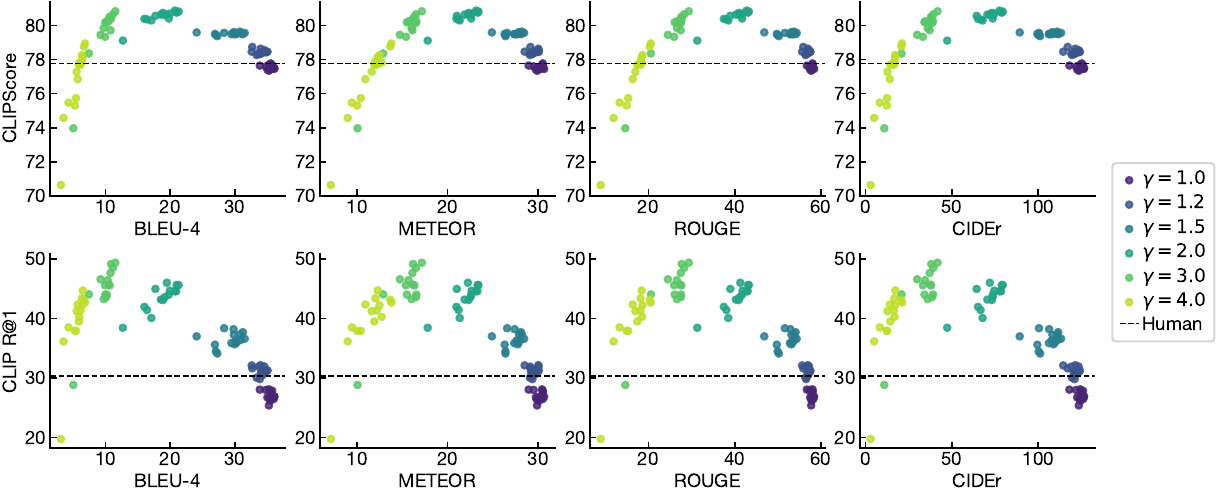}
\end{center}
\vspace{-0.5em}
\caption{Classifier-free guidance controls a trade off between reference-free and reference-based captioning metrics. Each point reflects a model trained with a different hyperparameter combination; each color represents a $\gamma$ value used to decode. Models are evaluated with different guidance scales $\gamma$, using reference-free captioning metrics based on CLIP ViT-B/32 (y-axes; top: CLIPScore, bottom: recall@1) and reference-based captioning metrics (x-axes). The dashed line reflects the value of the reference-free captioning metric for the ground-truth captions obtained from MS-COCO.}
\label{fig:clip_vs_cider}
\end{figure*}

\begin{table*}[t]
\begin{center}
\footnotesize
\setlength{\tabcolsep}{5pt}
\begin{tabular}{|l|rrrrl|lrrr|l|}
\hline
\multicolumn{1}{|c}{} & \multicolumn{5}{|c|}{Reference-Based Metrics} & \multicolumn{4}{|c|}{Reference-Free Metrics} &\\
\multicolumn{1}{|c|}{Model}  &\multicolumn{1}{c}{BLEU-4} &\multicolumn{1}{c}{METEOR} &\multicolumn{1}{c}{ROUGE} & \multicolumn{1}{c}{CIDEr} & \multicolumn{1}{c|}{RefOnlyCLIP-S} & \multicolumn{1}{c}{CLIP-S} &\multicolumn{1}{c}{R@1} &\multicolumn{1}{c}{R@5} &\multicolumn{1}{c|}{R@10}&\multicolumn{1}{c|}{RefCLIP-S}
\\ \hline \hline
\multicolumn{4}{|l}{\textit{Models trained with CLIP features or losses:}}&&&&&&&\\
CLIP-Captioner~\citep{barraco2022unreasonable} & 38.7 & 29.3 & 58.6 & 126.0 & 0.811 & 0.754 & & & & 0.814\\
UMT-BITG~\citep{huang2021unifying} & 37.3 & 28.2 & 57.9 & 122.6 & & 0.772 & & & &\\
X-LAN+SCST+GEG~\cite{zhang2022distincive} & 36.5 & 28.7 &57.5 & 121.7 & & & 28.1 & 50.3 & 67.2 &\\
CIDEr + CLIP-S Reward~\citep{cho2022fine} & 37.7 & 28.8 & 58.3 & 124.6 & & 0.772 & 24.4 & 50.2 & 63.1 & \\
CLIP-S Reward~\citep{cho2022fine} & 6.2 & 18.7 &31.6 & 11.2 & & 0.860 & 42.5 & 71.6 & 82.2 & \\
ZeroCap~\citep{tewel2022zerocap} & 2.6 & 11.5 & & 14.6 & & 0.87 &&& &0.79\\
\hline
\multicolumn{4}{|l}{\textit{Models trained without access to CLIP:}}&&&&&&&\\
UMT-BITG w/o CLIP loss~\citep{huang2021unifying} & 37.6 & 28.3 & 58.1 & 122.5 & & 0.725 & & & &\\
VinVL-large~\citep{zhang2021vinvl} &\textbf{41.0} & \textbf{30.9} & \textbf{59.4}$^*$ & \textbf{140.9} & \textbf{0.91}$^*$ & 0.78$^*$ & & & & \textbf{0.84}$^*$\\
\hline

          Ours ($\gamma = 1.0$) &    36.1 & 30.5 &     58.2 &  126.1 &                        0.900 &                0.775 &              26.5 &              51.9 &               64.1 &                    0.830 \\
          Ours ($\gamma = 1.2$) &    35.1 & 30.0 &     57.5 &  124.1 &                        0.899 &                0.785 &              31.3 &              57.4 &               69.3 &                    0.835 \\
          Ours ($\gamma = 1.5$) &    31.5 & 28.4 &     54.4 &  113.2 &                        0.891 &                0.796 &              36.6 &              64.0 &               75.0 &                    \textbf{0.838} \\
          Ours ($\gamma = 2.0$) &    20.9 & 23.3 &     43.0 &   78.6 &                        0.862 &                \textbf{0.808} &              44.6 &              71.7 &               81.7 &                    0.831 \\
          Ours ($\gamma = 3.0$) &    11.5 & 17.1 &     29.4 &   41.7 &                        0.820 &                \textbf{0.808} &              \textbf{49.4} &              \textbf{75.7} &               \textbf{84.7} &                    0.811 \\
          Ours ($\gamma = 4.0$) &     6.5 & 12.3 &     18.4 &   17.3 &                        0.766 &                0.782 &              44.7 &              71.3 &               80.9 &                    0.771 \\
\hline
\end{tabular}
\end{center}
\caption{Quantitative comparison of our approach with results from previous work that reports CLIP-based metrics. For VinVL-large, $^*$ indicates metrics from \citet{kasai2021bidimensional}. %
}
\label{tab:results}
\end{table*}

We adopt the standard reference-based captioning metrics BLEU-4, METEOR, ROUGE, and CIDEr, as well as reference-free captioning metrics based on CLIP ViT-B/32~\citep{radford2021learning}. The first reference-free captioning metric is CLIPScore~\citep{hessel-etal-2021-clipscore}, which is defined as
$\verb|CLIP-S|(\vc, \vv) = 2.5 \cdot \max(\cos(\vc, \vv), 0)$
where $\vc$ and $\vv$ are the CLIP embeddings of the caption and image respectively. The second reference-free metric measures the accuracy with which we can retrieve an image from the generated caption within a given test split by taking the $k$ nearest neighbors of the caption in the CLIP embedding space. Because recall@$k$ for $k > 1$ is highly correlated with recall@1 (R@5: $r=0.99$, R@10: $r=0.98$), we plot only recall@1. We additionally report RefOnlyCLIP-S, a reference-based metric that uses the CLIP text encoder to compute the similarity of CLIP embeddings of the generated captions with embeddings of ground truth captions, and RefCLIP-S, which takes the average of the per-image harmonic means of CLIP-S and RefOnlyCLIP-S~\citep{hessel-etal-2021-clipscore}. Unless otherwise stated, all evaluation is performed on the MS-COCO Karpathy test split~\cite{karpathy2015deep}.%

\section{Results}

\subsection{Classifier-free guidance}
We first investigate the trade-off between reference-based and reference-free image captioning metrics as a function of guidance scale. Because different guidance scales and metrics could conceivably benefit from different fine-tuning hyperparameter combinations, we plot all results from our hyperparameter grid in Figure~\ref{fig:clip_vs_cider}. Although standard greedy decoding ($\gamma = 1.0$) produces the highest CIDEr, METEOR, ROUGE, and BLEU-4 scores, higher guidance weights consistently yield higher values of reference-free captioning metrics. In particular, $\gamma = 3.0$ offers both the best caption$\to$image recall and the best CLIPScore. %

\begin{figure*}[t!]
\begin{center}
\tablefont
\setlength{\tabcolsep}{0pt}
\RaggedRight
\begin{tabular}{p{0.142\linewidth}p{0.142\linewidth}p{0.142\linewidth}p{0.142\linewidth}p{0.142\linewidth}p{0.142\linewidth}p{0.142\linewidth}}
\hfil\includegraphics[width=0.8\linewidth]{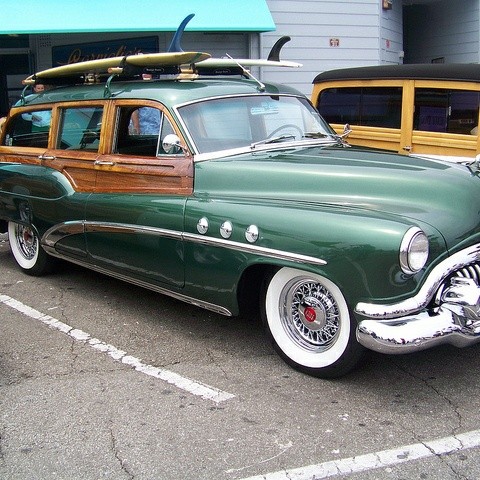} &
\hfil\includegraphics[width=0.8\linewidth]{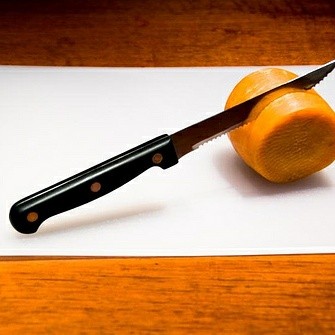} &
\hfil\includegraphics[width=0.8\linewidth]{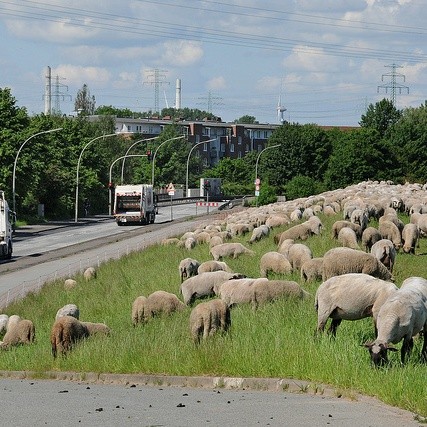} &
\hfil\includegraphics[width=0.8\linewidth]{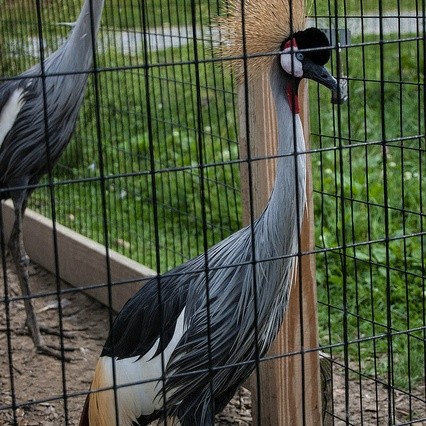} &
\hfil\includegraphics[width=0.8\linewidth]{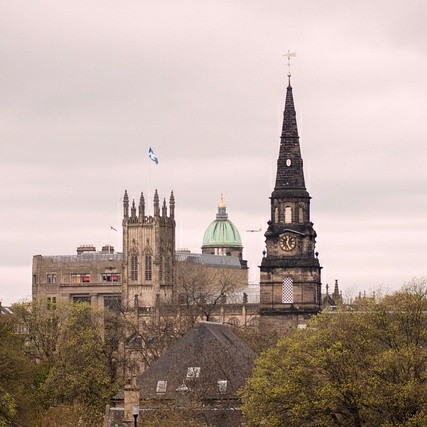} &
\hfil\includegraphics[width=0.8\linewidth]{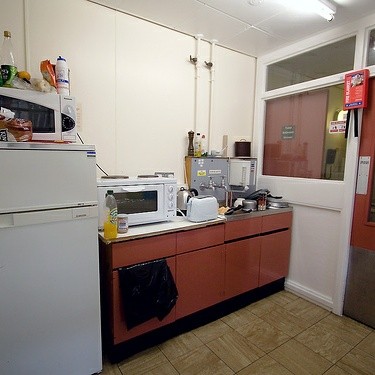} &
\hfil\includegraphics[width=0.8\linewidth]{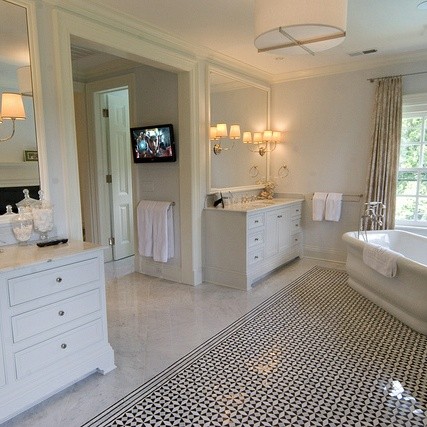}
\\{
\setlength{\tabcolsep}{1pt}
\begin{tabular}[t]{rp{0.74\linewidth}}
$\gamma$=1.0:& a car with a surfboard on top of it parked next to a car\\
$\gamma$=1.5:& a vintage station wagon with a surfboard on top\\
$\gamma$=2.0:& antique station wagons and a car buick stationwagon\\
$\gamma$=3.0:& buick woody woody stationwagon and surf green station wagon parked in front of car show establishment\\
GT: & An old green and brown car with chrome trim.
\end{tabular}
} &
{
\setlength{\tabcolsep}{1pt}
\begin{tabular}[t]{rp{0.74\linewidth}}
$\gamma$=1.0:& a knife is sitting on a cutting board next to a sliced carrot\\
$\gamma$=1.5:& a knife is sitting on a cutting board next to an orange\\
$\gamma$=2.0:& knife sitting on cutting board next to whole one\\
$\gamma$=3.0:& knife sitting on cutting board next to misshappi carrot on cutting board\\
GT:& A knife sticking out of the side of a block of cheese.
\end{tabular}
} &
{
\setlength{\tabcolsep}{1pt}
\begin{tabular}[t]{rp{0.74\linewidth}}
$\gamma$=1.0:& a herd of sheep grazing on a grass covered road\\
$\gamma$=1.5:& sheep grazing on a highway with a truck on the road\\
$\gamma$=2.0:& sheep graze on freeway medians where grass is grown\\
$\gamma$=3.0:& grazing trucks blocking sheep on roadway grazing grass\\
GT:& A large herd of sheep are grazing by the busy road.
\end{tabular}
} &
{
\setlength{\tabcolsep}{1pt}
\begin{tabular}[t]{rp{0.74\linewidth}}
$\gamma$=1.0:& two birds standing in a cage with their heads in the air\\
$\gamma$=1.5:& two birds standing inside of a cage in a zoo\\
$\gamma$=2.0:& two crested cranes inside a wire cage\\
$\gamma$=3.0:& crested tantalus cranes caged together in birdcage enclosure\\
GT:& Two birds who are looking out of the cage they are in.
\end{tabular}
} &
{
\setlength{\tabcolsep}{1pt}
\begin{tabular}[t]{rp{0.74\linewidth}}
$\gamma$=1.0:& a view of a city with a clock tower in the background\\
$\gamma$=1.5:& a city with steeples and trees and buildings\\
$\gamma$=2.0:& spires of churches line a city skyline\\
$\gamma$=3.0:& spires steeples buildings trees church spires and trees\\
GT:& A clock that is on the side of a tower.
\end{tabular}
} &
{
\setlength{\tabcolsep}{1pt}
\begin{tabular}[t]{rp{0.74\linewidth}}
$\gamma$=1.0:& a kitchen with a microwave and a refrigerator\\
$\gamma$=1.5:& a kitchen with a microwave and a refrigerator\\
$\gamma$=2.0:& a kitchen with red appliances and white cupboards\\
$\gamma$=3.0:& appliances sit in a small empty dingroomy red and white kitchen\\
GT:& A kitchen that has a tile floor, a refrigerator, a microwave, and a toaster.
\end{tabular}
} &
{
\setlength{\tabcolsep}{1pt}
\begin{tabular}[t]{rp{0.74\linewidth}}
$\gamma$=1.0:& a bathroom with a large mirror and a bathtub\\
$\gamma$=1.5:& a bathroom with a large mirror and a bathtub\\
$\gamma$=2.0:& a spacious bathroom with a large mirror and a large tub\\
$\gamma$=3.0:& spacious bathroom with chandelier over tub mirrors and tv\\
GT:& A bathroom with a tub, sinks, lights and a television.
\end{tabular}
}
\end{tabular}
\end{center}
\vspace{-0.5em}
\caption{Caption descriptiveness increases with CFG strength, but high CFG strengths produce agrammatical captions. Here we show examples of captions generated with different classifier-free guidance scales, for randomly selected images without human faces from the MS-COCO Karpathy test split. Captions labeled $\gamma=1.0$ are obtained without CFG; $\gamma > 1$ uses CFG; GT = ground truth.}
\vspace{-0.5em}
\label{fig:examples}
\end{figure*}

Table~\ref{tab:results} compares our results, obtained from a single model evaluated at different guidance scales, with previous work that reports either CLIPScore or CLIP ViT-B/32 caption$\to$image retrieval performance. Although our model is trained with standard cross-entropy loss rather than a CLIP-based loss and our pretraining dataset is distinct from CLIP's, sampling from our model with CFG yields higher CLIPScores than all other models trained without CLIP-based losses, and better CLIP caption$\to$image retrieval even when compared with models that use CLIP-based losses.

\begin{figure}[t]
    \centering
    \includegraphics[width=\linewidth]{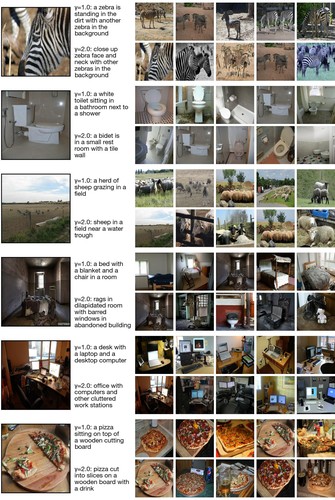}
    \vskip -0.15em
    \caption{Captions generated with CFG contain specific details that improve retrieval. For each reference image (far left), we show captions at $\gamma = 1.0$ (no guidance) and $\gamma = 2.0$. To the right, we show the closest images to each caption in the CoCa embedding space. Reference images are selected at random subject to the constraints that the closest image differs between $\gamma$ values and there are no identifiable human faces.
    }
    \vspace{-0.1em}
    \label{fig:retrieval}
\end{figure}

We present examples of captions generated at different CFG scales in Figure~\ref{fig:examples}. Higher CFG strengths lead to more descriptive captions. At $\gamma=1.0$, the central object in the top left image is described as a ``car'' as in the ground truth caption, whereas at $\gamma>1.0$ it is a ``station wagon.'' Similarly, at low CFG strengths, the birds in the center image are described simply as ``birds,'' whereas at $\gamma = 2.0$ they become ``crested cranes.'' However, at $\gamma = 3.0$, captions clearly become less grammatical, containing repeated words (``woody woody'') and nonsense words (``misshappi'', ``dingroomy''). Figure~\ref{fig:retrieval} shows captions obtained with and without CFG next to the top 5 closest images in the embedding space of CoCa 2B~\citet{yu2022coca},\footnote{We use CoCa 2B rather than CLIP because, quantitatively and qualitatively, it provides better retrieval results both with and without guidance.} where it is clear that CFG adds details to captions that help to distinguish them from other captions in the test split. We provide additional examples in Appendix~\ref{app:examples}. %

To provide additional quantitative assessments of the specificity of elicited captions, we perform two additional evaluations, described further in Appendix~\ref{app:quantitative_specificity}. First, we generate captions for the Stanford Dogs~\cite{KhoslaYaoJayadevaprakashFeiFei_FGVC2011} test set, which consists of 8,580 images in total of 120 breeds of dogs, and examine their properties. Without guidance, only 1.9\% of captions contain one of the 120 breed names, whereas at $\gamma = 2.0$, 42.4\% do. The percentage of these breed names that are correct changes little, from 61.7\% without guidance to 58.5\% at $\gamma=2.0$. Second, we performed a human evaluation comparing captions of MS-COCO test set images obtained without guidance and at $\gamma = 2.0$. We asked subjects to select the caption that is ``better'' and ``more descriptive'' or to indicate that they are both equal. When we asked these questions separately, we found that the two sets of captions are statistically indistinguishable. However, when asking both questions on the same survey, we found that captions generated without guidance are slightly ``better'' (50.5\% vs. 46.6\%, $p = 0.006$, binomial test) but captions generated at $\gamma = 2.0$ are ``more descriptive'' (52.7\% vs. 45.8\%, $p = 1 \times 10^{-6}$).

\begin{figure}
    \centering
    \includegraphics[width=\linewidth]{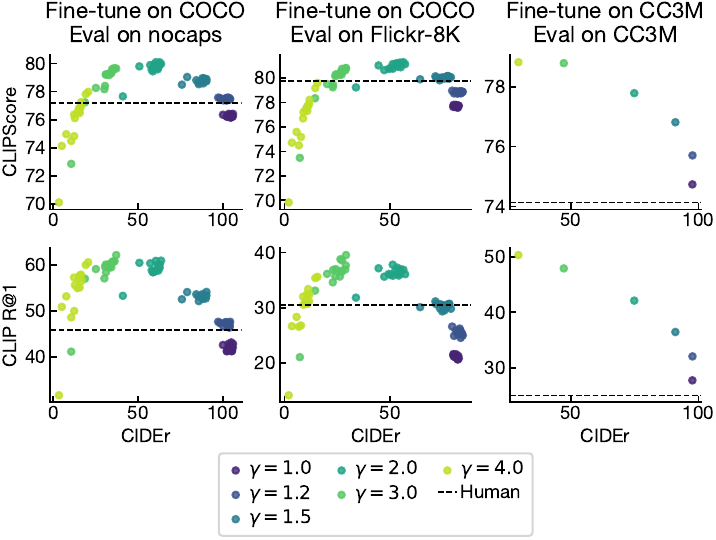}
    \caption{CFG also yields trade-offs between captioning metrics on nocaps, Flickr-8K, and CC3M.}
    \label{fig:nocaps}
    \vspace{-3em}
\end{figure}

To validate the reliability of our results, we further measure the impact of CFG on three additional datasets, \texttt{\textbf{nocaps}}~\cite{agrawal2019nocaps}, Flickr-8k~\cite{hodosh2013framing}, and Conceptual Captions (CC3M)~\cite{sharma2018conceptual}, as well as with alternative retrieval models. \texttt{\textbf{nocaps}} is a test set for captioning models with objects not present in MS-COCO; Flickr-8k is a small captioning dataset collected using a different procedure than MS-COCO; and Conceptual Captions is a set of 3.3M captions collected from filtered alt-text. We fine-tune the bottleneck CoCa-Base model directly on CC3M, and use our model fine-tuned on MS-COCO to caption images on \texttt{\textbf{nocaps}} and Flickr-8K. As shown in Figure~\ref{fig:nocaps}, we find trade-offs between reference-based and reference-free captioning metrics similar to those above. %
In Appendix~\ref{app:more_retrieval_models}, we report reference-free captioning metrics on MS-COCO computed with two additional retrieval models: the pretrained CoCa 2B model from \citet{yu2022coca} and the fine-tuned CoCa Base model that we use to generate captions. With both models, CFG substantially increases recall, in line with results obtained with CLIP ViT-B/32.

Although CFG produces captions that are more successful at uniquely identifying images than decoding from the conditional distribution, caption lengths are similar for $\gamma \in [1, 2]$, as shown in Table~\ref{tab:caption_length}. Thus, at low guidance strengths, CFG improves recall by making more efficient use of words, rather than by producing more verbose captions. Higher CFG strengths lead to longer captions but, as described above, these captions are agrammatical and contain nonsense words.

\begin{table}
\begin{center}
\footnotesize
\begin{tabular}{|l|r|r|}
\hline
$\gamma$ & Words & Characters \\
\hline\hline
$1.0$ & $9.6 \pm 1.4$ & $44.2 \pm 7.2$ \\
$1.2$ & $9.6 \pm 1.4$ & $44.7 \pm 7.4$ \\
$1.5$ & $9.4 \pm 1.4$ & $45.7 \pm 7.8$ \\
$2.0$ & $9.3 \pm 2.4$ & $50.3 \pm 18.6$ \\
$3.0$ & $10.7 \pm 7.6$ & $69.0 \pm 56.1$ \\
$4.0$ & $19.9 \pm 16.9$ & $161.2 \pm 140.0$ \\
\hline
\end{tabular}
\end{center}
\caption{Moderate CFG scales do not substantially change caption lengths, although higher CFG scales result in longer captions. Numbers are mean $\pm$ standard deviation.}
\label{tab:caption_length}
\vspace{-0.5em}
\end{table}

\subsection{Language model guidance}
\label{sec:lm_guidance}

We first experiment with guiding a captioning model fine-tuned on MS-COCO to produce more  descriptive captions using a language model prompted with manually written prompts. We first manually wrote a prompt containing 10 descriptive captions of COCO test set images (Appendix~\ref{app:captions}). We then sweep over $\alpha \in \{1, 2, 3, 4, 5, 6, 7, 8, 9, 10, 12, 15\}$ and $\beta \in \{0, \alpha/4, \alpha/2, 3/4\alpha, \alpha\}$, and compare the resulting retrieval/CIDEr trade-off to that produced by the same model with CFG. We observe that it is possible to obtain small improvements upon the Pareto frontier provided by CFG, as shown in
\begin{figure}
    \centering
    \includegraphics[width=0.8\linewidth]{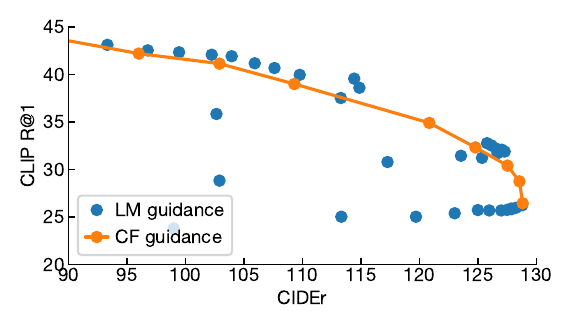}
    \vspace{-0.5em}
    \caption{Language model guidance produces captions that slightly exceed the Pareto frontier of CIDEr vs. caption$\to$image retrieval accuracy on MS-COCO.}
    \label{fig:pareto_improvement}
    \vspace{-0.5em}
\end{figure} Figure~\ref{fig:pareto_improvement}. With $\alpha = 5$, $\beta = -5/2$, LM guidance achieves CLIP ViT-B/32 R@1 of 39.6\% and CIDEr of 114.4, whereas CFG with $\gamma = 1.6$ is worse on both metrics, achieving R@1 of 39.0\% and CIDEr of 109.3.

\begin{figure}[t!]
\begin{center}
\tablefont
\setlength{\tabcolsep}{2pt}
\RaggedRight
\begin{tabular}{p{0.24\linewidth}p{0.24\linewidth}p{0.24\linewidth}p{0.24\linewidth}}
\hfil\includegraphics[width=1.0\linewidth]{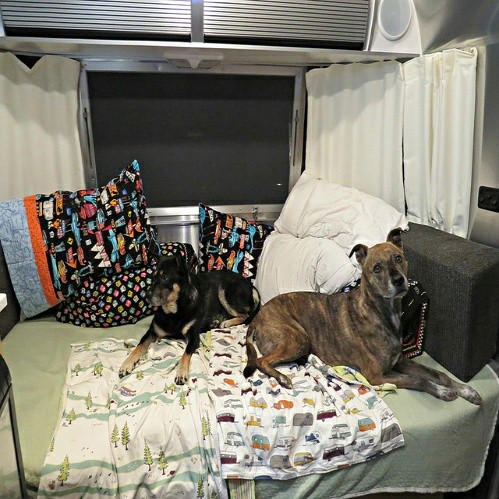} &
\hfil\includegraphics[width=1.0\linewidth]{figures/example_counting/1.jpg} &
\hfil\includegraphics[width=1.0\linewidth]{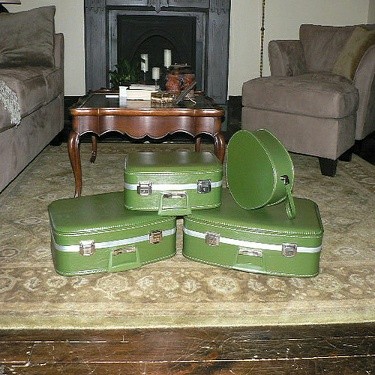} &
\hfil\includegraphics[width=1.0\linewidth]{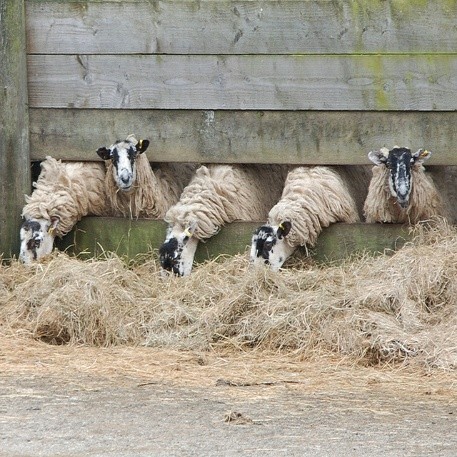}
\\ \centering a photo of two dogs &
 \centering a photo of one bird &
 \centering a photo of four suitcases &
 \centering a photo of five sheep
\end{tabular}
\end{center}
\vspace{-0.5em}
\caption{Captions generated with LM guidance with a prompt of 25 captions in the form of ``a photo of \texttt{NUMBER} \texttt{OBJECTS}''. Examples are selected to show different numbers of objects.}
\label{fig:counting}
\end{figure}

We further experiment with using prompting to control the captioner using a manually written prompt of 25 captions in the form of ``a photo of \verb!NUMBER! \verb!OBJECTS!'' (e.g., ``a photo of eight apples''; see Appendix~\ref{app:captions}). With $\alpha = \beta = 1$, the guided model is able to match this format and counts the number of objects in images (Figure~\ref{fig:counting}).

\begin{table*}[]
    \centering
    \footnotesize
\begin{tabular}{|l|rrrrr|rrrr|r|}
\hline
\multicolumn{1}{|c}{} & \multicolumn{5}{|c|}{Reference-Based Metrics} & \multicolumn{4}{|c|}{Reference-Free Metrics} &\\
\multicolumn{1}{|c|}{Model}  &\multicolumn{1}{c}{BLEU-4}  &\multicolumn{1}{c}{METEOR} &\multicolumn{1}{c}{ROUGE} & \multicolumn{1}{c}{CIDEr} & \multicolumn{1}{c|}{RefOnlyCLIP-S} & \multicolumn{1}{c}{CLIP-S} &\multicolumn{1}{c}{R@1} &\multicolumn{1}{c}{R@5} &\multicolumn{1}{c|}{R@10}&\multicolumn{1}{c|}{RefCLIP-S}
\\ \hline \hline
\multicolumn{4}{|l}{\textit{Classifier-free guidance:}}&&&&&&&\\
 $\gamma = 1.0$ &     8.2 &     8.3 &     21.8 &   21.8 &                        0.766 &                0.694 &               9.0 &              19.5 &               26.1 &                    0.725 \\
 $\gamma = 1.2$ &     8.6 &     9.5 &     24.5 &   25.0 &                        0.781 &                0.718 &              12.7 &              27.2 &               35.1 &                    0.745 \\
 $\gamma = 1.5$ &     8.9 &    10.0 &     25.6 &   25.2 &                        0.780 &                0.728 &              16.7 &              33.8 &               43.0 &                    0.750 \\
 $\gamma = 2.0$ &     8.1 &     9.7 &     23.9 &   22.9 &                        0.777 &                0.741 &              21.2 &              40.8 &               51.1 &                    0.755 \\
 $\gamma = 3.0$ &     7.1 &     8.7 &     20.0 &   18.5 &                        0.767 &                \textbf{0.753} &              25.8 &              47.8 &               58.3 &                    0.756 \\
 $\gamma = 4.0$ &     6.4 &     7.5 &     16.3 &   13.9 &                        0.749 &                0.743 &              \textbf{27.3} &              \textbf{48.5} &               \textbf{58.1} &                    0.742 \\
 \hline
\multicolumn{4}{|l}{\textit{Language model guidance with $\alpha=\beta=1$:}}&&&&&&&\\
       2 captions &    12.7 &    14.6 &     34.7 &   39.3 &                        0.806 &                0.688 &              10.0 &              23.7 &               32.4 &                    0.740 \\
       5 captions &    15.0 &    16.6 &     39.1 &   48.6 &                        0.827 &                0.712 &              12.4 &              27.5 &               37.5 &                    0.763 \\
      10 captions &    16.2 &    17.7 &     40.5 &   53.1 &                        0.835 &                0.723 &              13.0 &              30.5 &               41.0 &                    0.773 \\
      20 captions &    17.4 &    18.4 &     41.6 &   57.4 &                        0.839 &                0.728 &              14.4 &              32.2 &               42.7 &                    0.777 \\
      50 captions &    \textbf{18.1} &    \textbf{19.1} &     \textbf{42.5} &   \textbf{59.7} &                        \textbf{0.840} &                0.729 &              13.4 &              32.5 &               43.9 &                    \textbf{0.778}\\
\hline\hline
\multicolumn{6}{|l|}{\textit{Other models trained without aligned MS-COCO images and captions:}}&&&&&\\
ZeroCap~\citep{tewel2022zerocap} & 2.6 & 11.5 && 14.6 && 0.87 &&& &0.79\\
MAGIC~\cite{su2022language} & 12.9 & 17.4 & 39.9 & 49.3 & &&&&&\\
Flamingo~\cite{alayrac2022flamingo} & &&& 84.3 & &&&&&\\
DeCap (560 captions)~\cite{lidecap} & & & & 51.4 &&&&&&\\
DeCap (full train set)~\cite{lidecap} & 24.7 & 25.0 & & 91.2 &&&&&&\\
CapDec (full train set)~\cite{nukrai2022text} & 26.4 & 25.1 & 51.8 & 91.8 &&&&&&\\
\hline
\end{tabular}
\vspace{0.5em}
    \caption{Comparison of decoding strategies for a captioning model trained only on minimally curated web data (JFT-5B and ALIGN) and evaluated on MS-COCO. At the bottom, we report metrics for other models trained without aligned MS-COCO images and captions. These models may not be directly comparable since they use different pretraining data. DeCap and CapDec use all 560K captions in the MS-COCO training set to train their decoders; we include CIDEr for DeCap with 560 captions (0.1\% of the training data) for comparison.}
    \label{tab:zero_shot}
\vspace{-0.7em}
\end{table*}

We next investigate whether language model guidance can elicit better captions from a model trained only on low-quality data. Here, we use a CoCa model that is pretrained on image-alt text pairs from the web (the ALIGN dataset~\cite{jia2021scaling}) and classification labels converted to text (the JFT-5B dataset~\cite{zhai2022scaling}), without any additional fine-tuning. Because the data distribution places higher probability mass on short, non-descriptive captions than on longer captions, the resulting model is of limited utility for captioning, and would generally need to be fine-tuned on another dataset such as MS-COCO before being applied to a captioning task. Rather than fine-tune, we use LM guidance to prompt the model with captions from the MS-COCO training set.

\begin{figure}
    \centering
    \includegraphics[width=\linewidth]{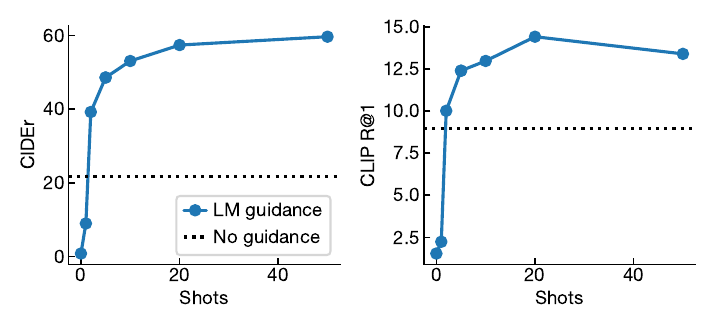}
    \vspace{-1.6em}
    \caption{LM guidance substantially improves CIDEr and retrieval scores of a model trained solely on minimally curated web data and evaluated on MS-COCO. The x-axis shows the number of captions used to prompt the LM; we do not prompt with images.}
    \label{fig:zero_shot_lm}
    \vspace{-1em}
\end{figure}

LM guidance substantially improves the quality of the captions produced by the original pretrained CoCa model without any clean parallel data. With LM guidance, we achieve CIDEr scores of 48.6 with 5 shots and 59.7 with 50 shots, far exceeding the CIDEr score of 21.8 obtained with no guidance. Figure~\ref{fig:zero_shot_lm} shows CIDEr and CLIP recall@1 scores for LM guidance of this pretrained CoCa model as a function of the number of shots, with $\alpha = \beta = 1$. Table~\ref{tab:zero_shot} compares classifier-free guidance and LM guidance. CFG yields higher CLIP-Scores and retrieval accuracy than LM guidance with $\alpha = \beta = 1$, but LM guidance provides much higher CIDEr scores.

\begin{figure}[t!]
\begin{center}
\tablefont
\setlength{\tabcolsep}{1pt}
\RaggedRight
\begin{tabular}{p{0.242\linewidth}p{0.242\linewidth}p{0.242\linewidth}p{0.242\linewidth}}
\hfil\includegraphics[width=1.0\linewidth]{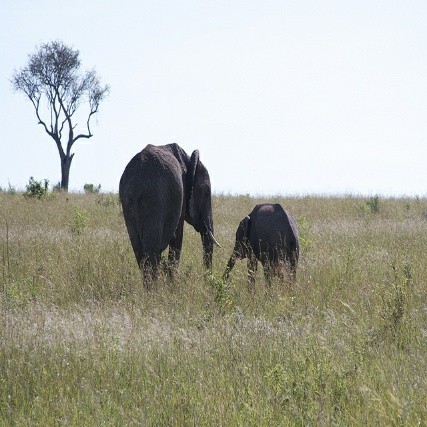} &
\hfil\includegraphics[width=1.0\linewidth]{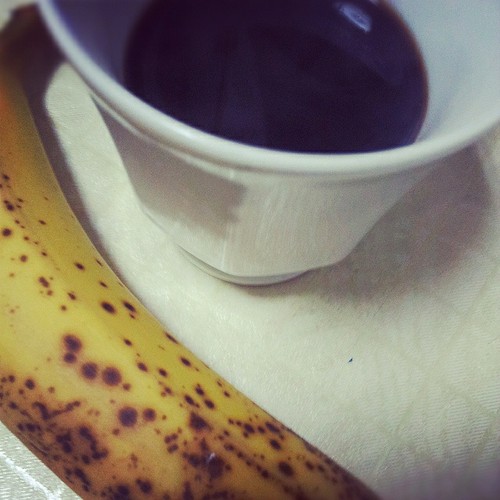} &
\hfil\includegraphics[width=1.0\linewidth]{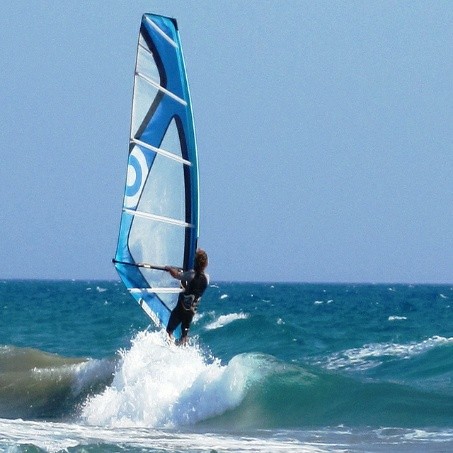} &
\hfil\includegraphics[width=1.0\linewidth]{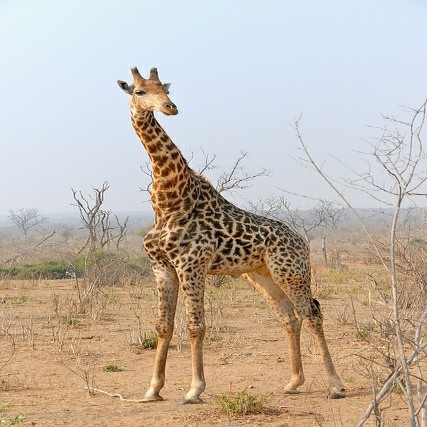}
\\{
\setlength{\tabcolsep}{1pt}
\begin{tabular}[t]{rp{0.68\linewidth}}
$\gamma$=1.0:& a photo of the small elephant.\\
$\gamma$=2.0:& elephants in the ruaha national park\\
$\gamma$=3.0:& elephants chobe np\\
LM:& a elephant and a zebra in a field\\
GT:& Two elephants standing on a grassy field next to a tree.
\end{tabular}
} &
{
\setlength{\tabcolsep}{1pt}
\begin{tabular}[t]{rp{0.68\linewidth}}
$\gamma$=1.0:& a photo of the small coffee.\\
$\gamma$=2.0:& a coffee in a video game.\\
$\gamma$=3.0:& a banana in a video game.\\
LM:& a banana with a cup of coffee\\
GT:& A close up of a banana next to a cup with liquid.
\end{tabular}
} &
{
\setlength{\tabcolsep}{1pt}
\begin{tabular}[t]{rp{0.68\linewidth}}
$\gamma$=1.0:& a photo of the large windsurfing.\\
$\gamma$=2.0:& windsurfing in tarifa\\
$\gamma$=3.0:& windsurfing wallpapers 1200x1024\\
LM:& a windsurfer in the water\\
GT:& A man riding a wind sail in the ocean filled with waves.
\end{tabular}
} &
{
\setlength{\tabcolsep}{1pt}
\begin{tabular}[t]{rp{0.68\linewidth}}
$\gamma$=1.0:& a photo of the large giraffe.\\
$\gamma$=2.0:& a giraffe in a video game.\\
$\gamma$=3.0:& giraffe standing photo 1\\
LM:& a giraffe standing in a tall tree\\
GT:& A giraffe in a dry savannah with dry shrubs
\end{tabular}
}
\end{tabular}
\end{center}
\vspace{-0.3em}
\caption{Examples of captions generated from a model pretrained only on minimally curated data, for randomly selected images without human faces. Captions labeled $\gamma=1.0$ are obtained without CFG; $\gamma > 1$ uses CFG; LM indicates LM guidance with $\alpha = \beta = 1$ and 20 shots; GT indicates ground truth.}
\label{fig:examples_zero_shot}
\vspace{-0.7em}
\end{figure}

We compare captions generated with CFG to those generated with LM guidance for four images in Figure~\ref{fig:examples_zero_shot}. In general, CFG produces agrammatical captions, whereas LM guidance produces grammatical captions but hallucinates details. For example, the image in the upper left shows two elephants and no zebras, but LM guidance leads to the caption ``an elephant and a zebra in a field.''

\section{Conclusion}
Our study indicates that it is possible to substantially improve the extent to which generated captions uniquely describe the images they correspond to, raising questions regarding the goal of image captioning and how it should be evaluated. As it is conventionally formulated, image captioning aims not to provide text that can substitute for an image, but to write the text that a human annotator would have written. This formulation penalizes captions that are more descriptive than ground truth, even when a human might prefer them. On the other hand, treating image captioning as a problem of generating a caption that lies close to the image in the embedding space of an image-text model is also inadequate, because captions that lie close to the image need not be grammatical and may contain gibberish. Our proposed methods leveraging classifier-free guidance and language model guidance modulate the trade-offs between these two goals, as captured by various reference-based and reference-free metrics.

There are several possible extensions to our work. First, our present experiments use only greedy decoding. Although greedy decoding appears to perform reasonably well in our setup, it may be suboptimal for LM guidance with prompts that impose greater structure upon the captions. If the LM is prompted to output either ``there is a person in this image'' or ``there is no person this image'', greedy decoding is likely to fail even if the captioner properly scores the two possible captions, because when choosing between the tokens ``a'' and ``no'', the captioner has no knowledge of the structure that the LM will impose on future tokens. Since beam search could explore both tokens, it may offer better results in this scenario. Second, our method could be combined with RL-based methods to increase similarity in a contrastive embedding space, which may further improve retrieval performance and CLIPScore. Finally, with a perfect captioning model, $p(\text{image}|\text{caption})$ should increase with $\gamma$. However, in practice we find that $\gamma > 3$ leads to a decrease in retrieval performance. This discrepancy suggests that the difference between the conditional and unconditional model distributions may be a noisy estimator of the pointwise mutual information. Although selecting $\gamma$ is one way to regularize this estimator, there may also be strategies to regularize $p(\rx|\ry)/p(\rx)$ at training time.

\section*{Acknowledgements}

\noindent We thank Kevin Clark, David Fleet, Geoffrey Hinton, and the rest of the Google DeepMind Toronto team for inspiration, comments, and discussion.

{\small
\bibliographystyle{ieee_fullname}
\bibliography{main}

\begin{thebibliography}{10}\itemsep=-1pt

\bibitem{aghajanyan2022cm3}
Armen Aghajanyan, Bernie Huang, Candace Ross, Vladimir Karpukhin, Hu Xu, Naman
  Goyal, Dmytro Okhonko, Mandar Joshi, Gargi Ghosh, Mike Lewis, et~al.
\newblock Cm3: A causal masked multimodal model of the internet.
\newblock {\em arXiv preprint arXiv:2201.07520}, 2022.

\bibitem{agrawal2019nocaps}
Harsh Agrawal, Karan Desai, Yufei Wang, Xinlei Chen, Rishabh Jain, Mark
  Johnson, Dhruv Batra, Devi Parikh, Stefan Lee, and Peter Anderson.
\newblock nocaps: novel object captioning at scale.
\newblock In {\em Proceedings of the IEEE International Conference on Computer
  Vision}, pages 8948--8957, 2019.

\bibitem{alayrac2022flamingo}
Jean-Baptiste Alayrac, Jeff Donahue, Pauline Luc, Antoine Miech, Iain Barr,
  Yana Hasson, Karel Lenc, Arthur Mensch, Katherine Millican, Malcolm Reynolds,
  et~al.
\newblock Flamingo: a visual language model for few-shot learning.
\newblock In {\em Advances in Neural Information Processing Systems}, 2022.

\bibitem{barraco2022unreasonable}
Manuele Barraco, Marcella Cornia, Silvia Cascianelli, Lorenzo Baraldi, and Rita
  Cucchiara.
\newblock The unreasonable effectiveness of clip features for image captioning:
  An experimental analysis.
\newblock In {\em Proceedings of the IEEE/CVF Conference on Computer Vision and
  Pattern Recognition}, pages 4662--4670, 2022.

\bibitem{chan2023ic}
David~M Chan, Austin Myers, Sudheendra Vijayanarasimhan, David~A Ross, and John
  Canny.
\newblock $ ic^3$: Image captioning by committee consensus.
\newblock {\em arXiv preprint arXiv:2302.01328}, 2023.

\bibitem{cho2022fine}
Jaemin Cho, Seunghyun Yoon, Ajinkya Kale, Franck Dernoncourt, Trung Bui, and
  Mohit Bansal.
\newblock Fine-grained image captioning with {CLIP} reward.
\newblock In {\em Findings of the Association for Computational Linguistics:
  NAACL 2022}, pages 517--527, Seattle, United States, July 2022. Association
  for Computational Linguistics.

\bibitem{chowdhery2022palm}
Aakanksha Chowdhery, Sharan Narang, Jacob Devlin, Maarten Bosma, Gaurav Mishra,
  Adam Roberts, Paul Barham, Hyung~Won Chung, Charles Sutton, Sebastian
  Gehrmann, et~al.
\newblock {PaLM}: Scaling language modeling with pathways.
\newblock {\em arXiv preprint arXiv:2204.02311}, 2022.

\bibitem{crowson2022classifierfree}
Katherine Crowson.
\newblock You can apply a similar trick to classifier-free guidance to
  autoregressive transformers to sample from a synthetic "super-conditioned"
  distribution.
\newblock {\footnotesize
  https://twitter.com/RiversHaveWings/status/1478093658716966912}, 2022.

\bibitem{daille1994approche}
B{\'e}atrice Daille.
\newblock {\em Approche mixte pour l’extraction automatique de terminologie:
  statistiques lexicales et filtres linguistiques}.
\newblock PhD thesis, Ph. D. thesis, Universit{\'e} Paris 7, 1994.

\bibitem{devlin-etal-2019-bert}
Jacob Devlin, Ming-Wei Chang, Kenton Lee, and Kristina Toutanova.
\newblock {BERT}: Pre-training of deep bidirectional transformers for language
  understanding.
\newblock In {\em Proceedings of the 2019 Conference of the North {A}merican
  Chapter of the Association for Computational Linguistics: Human Language
  Technologies, Volume 1 (Long and Short Papers)}, pages 4171--4186,
  Minneapolis, Minnesota, June 2019. Association for Computational Linguistics.

\bibitem{devlin2015language}
Jacob Devlin, Hao Cheng, Hao Fang, Saurabh Gupta, Li Deng, Xiaodong He,
  Geoffrey Zweig, and Margaret Mitchell.
\newblock Language models for image captioning: The quirks and what works.
\newblock {\em arXiv preprint arXiv:1505.01809}, 2015.

\bibitem{gafni2022make}
Oran Gafni, Adam Polyak, Oron Ashual, Shelly Sheynin, Devi Parikh, and Yaniv
  Taigman.
\newblock Make-a-scene: Scene-based text-to-image generation with human priors.
\newblock {\em arXiv preprint arXiv:2203.13131}, 2022.

\bibitem{hessel-etal-2021-clipscore}
Jack Hessel, Ari Holtzman, Maxwell Forbes, Ronan Le~Bras, and Yejin Choi.
\newblock {CLIPS}core: A reference-free evaluation metric for image captioning.
\newblock In {\em Proceedings of the 2021 Conference on Empirical Methods in
  Natural Language Processing}, pages 7514--7528, Online and Punta Cana,
  Dominican Republic, Nov. 2021. Association for Computational Linguistics.

\bibitem{ho2021classifierfree}
Jonathan Ho and Tim Salimans.
\newblock Classifier-free diffusion guidance.
\newblock In {\em NeurIPS 2021 Workshop on Deep Generative Models and
  Downstream Applications}, 2021.

\bibitem{hodosh2013framing}
Micah Hodosh, Peter Young, and Julia Hockenmaier.
\newblock Framing image description as a ranking task: Data, models and
  evaluation metrics.
\newblock {\em Journal of Artificial Intelligence Research}, 47:853--899, 2013.

\bibitem{huang2021unifying}
Yupan Huang, Hongwei Xue, Bei Liu, and Yutong Lu.
\newblock Unifying multimodal transformer for bi-directional image and text
  generation.
\newblock In {\em Proceedings of the 29th ACM International Conference on
  Multimedia}, pages 1138--1147, 2021.

\bibitem{jia2021scaling}
Chao Jia, Yinfei Yang, Ye Xia, Yi-Ting Chen, Zarana Parekh, Hieu Pham, Quoc Le,
  Yun-Hsuan Sung, Zhen Li, and Tom Duerig.
\newblock Scaling up visual and vision-language representation learning with
  noisy text supervision.
\newblock In {\em International Conference on Machine Learning}, pages
  4904--4916. PMLR, 2021.

\bibitem{karpathy2015deep}
Andrej Karpathy and Fei{-}Fei Li.
\newblock Deep visual-semantic alignments for generating image descriptions.
\newblock In {\em {IEEE} Conference on Computer Vision and Pattern Recognition,
  {CVPR} 2015, Boston, MA, USA, June 7-12, 2015}, pages 3128--3137. {IEEE}
  Computer Society, 2015.

\bibitem{kasai2021bidimensional}
Jungo Kasai, Keisuke Sakaguchi, Ronan~Le Bras, Lavinia Dunagan, Jacob Morrison,
  Alexander~R Fabbri, Yejin Choi, and Noah~A Smith.
\newblock Bidimensional leaderboards: Generate and evaluate language hand in
  hand.
\newblock {\em arXiv preprint arXiv:2112.04139}, 2021.

\bibitem{kasai2021transparent}
Jungo Kasai, Keisuke Sakaguchi, Lavinia Dunagan, Jacob Morrison, Ronan~Le Bras,
  Yejin Choi, and Noah~A Smith.
\newblock Transparent human evaluation for image captioning.
\newblock {\em arXiv preprint arXiv:2111.08940}, 2021.

\bibitem{KhoslaYaoJayadevaprakashFeiFei_FGVC2011}
Aditya Khosla, Nityananda Jayadevaprakash, Bangpeng Yao, and Li Fei-Fei.
\newblock Novel dataset for fine-grained image categorization.
\newblock In {\em First Workshop on Fine-Grained Visual Categorization, IEEE
  Conference on Computer Vision and Pattern Recognition}, Colorado Springs, CO,
  June 2011.

\bibitem{lidecap}
Wei Li, Linchao Zhu, Longyin Wen, and Yi Yang.
\newblock Decap: Decoding clip latents for zero-shot captioning.
\newblock In {\em International Conference on Learning Representations}, 2022.

\bibitem{lin2004rouge}
Chin-Yew Lin.
\newblock {ROUGE}: A package for automatic evaluation of summaries.
\newblock In {\em Text summarization branches out}, pages 74--81, 2004.

\bibitem{lin2014microsoft}
Tsung-Yi Lin, Michael Maire, Serge Belongie, James Hays, Pietro Perona, Deva
  Ramanan, Piotr Doll{\'a}r, and C~Lawrence Zitnick.
\newblock Microsoft coco: Common objects in context.
\newblock In {\em Computer Vision--ECCV 2014: 13th European Conference, Zurich,
  Switzerland, September 6-12, 2014, Proceedings, Part V 13}, pages 740--755.
  Springer, 2014.

\bibitem{lindh2018generating}
Annika Lindh, Robert~J Ross, Abhijit Mahalunkar, Giancarlo Salton, and John~D
  Kelleher.
\newblock Generating diverse and meaningful captions.
\newblock In {\em International Conference on Artificial Neural Networks},
  pages 176--187. Springer, 2018.

\bibitem{nichol2021glide}
Alexander~Quinn Nichol, Prafulla Dhariwal, Aditya Ramesh, Pranav Shyam, Pamela
  Mishkin, Bob Mcgrew, Ilya Sutskever, and Mark Chen.
\newblock Glide: Towards photorealistic image generation and editing with
  text-guided diffusion models.
\newblock In {\em International Conference on Machine Learning}, pages
  16784--16804. PMLR, 2022.

\bibitem{nukrai2022text}
David Nukrai, Ron Mokady, and Amir Globerson.
\newblock Text-only training for image captioning using noise-injected clip.
\newblock {\em arXiv preprint arXiv:2211.00575}, 2022.

\bibitem{papineni2002bleu}
Kishore Papineni, Salim Roukos, Todd Ward, and Wei-Jing Zhu.
\newblock {BLEU}: a method for automatic evaluation of machine translation.
\newblock In {\em Proceedings of the 40th annual meeting of the Association for
  Computational Linguistics}, pages 311--318, 2002.

\bibitem{pont2020connecting}
Jordi Pont-Tuset, Jasper Uijlings, Soravit Changpinyo, Radu Soricut, and
  Vittorio Ferrari.
\newblock Connecting vision and language with localized narratives.
\newblock In {\em Computer Vision--ECCV 2020: 16th European Conference,
  Glasgow, UK, August 23--28, 2020, Proceedings, Part V 16}, pages 647--664.
  Springer, 2020.

\bibitem{radford2021learning}
Alec Radford, Jong~Wook Kim, Chris Hallacy, Aditya Ramesh, Gabriel Goh,
  Sandhini Agarwal, Girish Sastry, Amanda Askell, Pamela Mishkin, Jack Clark,
  et~al.
\newblock Learning transferable visual models from natural language
  supervision.
\newblock In {\em International Conference on Machine Learning}, pages
  8748--8763. PMLR, 2021.

\bibitem{radford2018improving}
Alec Radford, Karthik Narasimhan, Tim Salimans, Ilya Sutskever, et~al.
\newblock Improving language understanding by generative pre-training.
\newblock 2018.

\bibitem{ramesh2022hierarchical}
Aditya Ramesh, Prafulla Dhariwal, Alex Nichol, Casey Chu, and Mark Chen.
\newblock Hierarchical text-conditional image generation with clip latents.
\newblock {\em arXiv preprint arXiv:2204.06125}, 2022.

\bibitem{rombach2022high}
Robin Rombach, Andreas Blattmann, Dominik Lorenz, Patrick Esser, and Bj{\"o}rn
  Ommer.
\newblock High-resolution image synthesis with latent diffusion models.
\newblock In {\em Proceedings of the IEEE/CVF Conference on Computer Vision and
  Pattern Recognition}, pages 10684--10695, 2022.

\bibitem{sahariaphotorealistic}
Chitwan Saharia, William Chan, Saurabh Saxena, Lala Li, Jay Whang, Emily
  Denton, Seyed Kamyar~Seyed Ghasemipour, Raphael Gontijo-Lopes, Burcu~Karagol
  Ayan, Tim Salimans, et~al.
\newblock Photorealistic text-to-image diffusion models with deep language
  understanding.
\newblock In {\em Advances in Neural Information Processing Systems}.

\bibitem{sanchez2023stay}
Guillaume Sanchez, Honglu Fan, Alexander Spangher, Elad Levi, Pawan~Sasanka
  Ammanamanchi, and Stella Biderman.
\newblock Stay on topic with classifier-free guidance.
\newblock {\em arXiv preprint arXiv:2306.17806}, 2023.

\bibitem{sharma2018conceptual}
Piyush Sharma, Nan Ding, Sebastian Goodman, and Radu Soricut.
\newblock Conceptual captions: A cleaned, hypernymed, image alt-text dataset
  for automatic image captioning.
\newblock In {\em Proceedings of the 56th Annual Meeting of the Association for
  Computational Linguistics (Volume 1: Long Papers)}, pages 2556--2565, 2018.

\bibitem{so2021primer}
David~R So, Wojciech Ma{\'n}ke, Hanxiao Liu, Zihang Dai, Noam Shazeer, and
  Quoc~V Le.
\newblock Primer: Searching for efficient transformers for language modeling.
\newblock {\em arXiv preprint arXiv:2109.08668}, 2021.

\bibitem{su2022language}
Yixuan Su, Tian Lan, Yahui Liu, Fangyu Liu, Dani Yogatama, Yan Wang, Lingpeng
  Kong, and Nigel Collier.
\newblock Language models can see: plugging visual controls in text generation.
\newblock {\em arXiv preprint arXiv:2205.02655}, 2022.

\bibitem{sun2017revisiting}
Chen Sun, Abhinav Shrivastava, Saurabh Singh, and Abhinav Gupta.
\newblock Revisiting unreasonable effectiveness of data in deep learning era.
\newblock In {\em Proceedings of the IEEE international conference on computer
  vision}, pages 843--852, 2017.

\bibitem{tewel2022zerocap}
Yoad Tewel, Yoav Shalev, Idan Schwartz, and Lior Wolf.
\newblock Zerocap: Zero-shot image-to-text generation for visual-semantic
  arithmetic.
\newblock In {\em Proceedings of the IEEE/CVF Conference on Computer Vision and
  Pattern Recognition}, pages 17918--17928, 2022.

\bibitem{vedantam2015cider}
Ramakrishna Vedantam, C Lawrence~Zitnick, and Devi Parikh.
\newblock {CIDEr}: Consensus-based image description evaluation.
\newblock In {\em Proceedings of the IEEE conference on computer vision and
  pattern recognition}, pages 4566--4575, 2015.

\bibitem{vinyals2015show}
Oriol Vinyals, Alexander Toshev, Samy Bengio, and Dumitru Erhan.
\newblock Show and tell: A neural image caption generator.
\newblock In {\em Proceedings of the IEEE conference on computer vision and
  pattern recognition}, pages 3156--3164, 2015.

\bibitem{wen2023hard}
Yuxin Wen, Neel Jain, John Kirchenbauer, Micah Goldblum, Jonas Geiping, and Tom
  Goldstein.
\newblock Hard prompts made easy: Gradient-based discrete optimization for
  prompt tuning and discovery.
\newblock {\em arXiv preprint arXiv:2302.03668}, 2023.

\bibitem{xu2022clip}
Shitong Xu.
\newblock Clip-diffusion-lm: Apply diffusion model on image captioning.
\newblock {\em arXiv preprint arXiv:2210.04559}, 2022.

\bibitem{yang2017dense}
Linjie Yang, Kevin Tang, Jianchao Yang, and Li-Jia Li.
\newblock Dense captioning with joint inference and visual context.
\newblock In {\em Proceedings of the IEEE conference on computer vision and
  pattern recognition}, pages 2193--2202, 2017.

\bibitem{yu2022coca}
Jiahui Yu, Zirui Wang, Vijay Vasudevan, Legg Yeung, Mojtaba Seyedhosseini, and
  Yonghui Wu.
\newblock Coca: Contrastive captioners are image-text foundation models.
\newblock {\em arXiv preprint arXiv:2205.01917}, 2022.

\bibitem{yu2022scaling}
Jiahui Yu, Yuanzhong Xu, Jing~Yu Koh, Thang Luong, Gunjan Baid, Zirui Wang,
  Vijay Vasudevan, Alexander Ku, Yinfei Yang, Burcu~Karagol Ayan, et~al.
\newblock Scaling autoregressive models for content-rich text-to-image
  generation.
\newblock {\em arXiv preprint arXiv:2206.10789}, 2022.

\bibitem{zhai2022scaling}
Xiaohua Zhai, Alexander Kolesnikov, Neil Houlsby, and Lucas Beyer.
\newblock Scaling vision transformers.
\newblock In {\em Proceedings of the IEEE/CVF Conference on Computer Vision and
  Pattern Recognition}, pages 12104--12113, 2022.

\bibitem{zhang2021vinvl}
Pengchuan Zhang, Xiujun Li, Xiaowei Hu, Jianwei Yang, Lei Zhang, Lijuan Wang,
  Yejin Choi, and Jianfeng Gao.
\newblock Vinvl: Revisiting visual representations in vision-language models.
\newblock In {\em Proceedings of the IEEE/CVF Conference on Computer Vision and
  Pattern Recognition}, pages 5579--5588, 2021.

\bibitem{zhang2022distincive}
Youyuan Zhang, Jiuniu Wang, Hao Wu, and Wenjia Xu.
\newblock Distincive image captioning via clip guided group optimization.
\newblock {\em arXiv preprint arXiv:2208.04254}, 2022.

\bibitem{zhu2022exploring}
Zixin Zhu, Yixuan Wei, Jianfeng Wang, Zhe Gan, Zheng Zhang, Le Wang, Gang Hua,
  Lijuan Wang, Zicheng Liu, and Han Hu.
\newblock Exploring discrete diffusion models for image captioning.
\newblock {\em arXiv preprint arXiv:2211.11694}, 2022.

\end{thebibliography}
}

\onecolumn
\counterwithin{figure}{section}
\counterwithin{table}{section}
\part*{Appendix}
\appendix

\section{Additional details regarding our approach}

\subsection{Pseudocode for classifier-free guidance}
Below, we provide pseudocode for greedy decoding with classifier-free guidance. Note that, in practice, we perform decoding in batches.
\label{app:pseudocode_cfg}
\begin{python}
# captioner: Captioning model (returns token log probs)
# img_embed: Image embedding
# gamma: Classifier-free guidance scale
# max_length: Maximum number of tokens in caption
# BOS: Beginning of sequence token
# EOS: End of sequence token

tokens = [BOS]
for i in range(0, max_length):
  # Eq. 3 (without the softmax, since it does not affect the argmax).
  cond_log_probs = captioner(tokens, img_embed)
  uncond_log_probs = captioner(tokens, zeros_like(img_embed))
  scores = uncond_log_probs + gamma * (cond_log_probs - uncond_log_probs)

  # Greedily take the next token.
  next_token = argmax(scores)
  tokens.append(next_token)
  if next_token == EOS: break
\end{python}

\subsection{Derivation of language model guidance}
\label{app:derivation_lmg}
\noindent Assume that we have two joint distributions of captions $\rx$ and images $\ry$, $p(\rx, \ry)$ and $q(\rx, \ry)$, and these distributions have the same pointwise mutual information between any image-caption pair, i.e. $\log \frac{q(\rx, \ry)}{q(\rx) q(\ry)} =  \log \frac{p(\rx, \ry)}{p(\rx) p(\ry)}$, and thus $\frac{q(\rx, \ry)}{q(\rx) q(\ry)} = \frac{p(\rx, \ry)}{p(\rx) p(\ry)}$. Starting with the leftmost expression from Eq. 2, there exists an expression that uses the joint distribution from $p$ but only marginals of captions from $q$,
\begin{align}
    \textstyle q(\rx)\left(\frac{q(\rx|\ry)}{q(\rx)}\right)^\gamma &= \textstyle q(\rx)\left(\frac{q(\rx,\ry)}{q(\rx)q(\ry)}\right)^\gamma\\
    &= \textstyle q(\rx)\left(\frac{p(\rx,\ry)}{p(\rx)p(\ry)}\right)^\gamma.
    \label{eq:lmg_shared_exp}
\end{align}

\noindent In Eq.~\ref{eq:lmg}, we further decouple the exponents for the numerator and denominator of the above equation. As we note, this decoupling is reminiscent of $\mathrm{pmi}^k$. To see this relationship, first note that $\frac{p(\rx, \ry)}{p(\rx) p(\ry)}$ is the exponential of $\mathrm{pmi}(\rx, \ry) = \log \frac{p(\rx, \ry)}{p(\rx) p(\ry)}$. Replacing $\mathrm{pmi}(\rx, \ry)$ with $\mathrm{pmi}^k(\rx, \ry) = \log \frac{p(x, y)^k}{p(x)p(y)}$, Eq.~\ref{eq:lmg_shared_exp} becomes $q(\rx)\left(\frac{p(\rx,\ry)^k}{p(\rx)p(\ry)}\right)^\gamma$.
Setting $\alpha = k\gamma$ and $\beta = \gamma$ gives
\begin{align}
    q(\rx)\left(\frac{p(\rx,\ry)^\alpha}{p(\rx)^\beta p(\ry)^\beta}\right) = q(\rx)\left(\frac{p(\rx|\ry)^\alpha}{p(\rx)^\beta p(\ry)^{\beta - \alpha}}\right) \propto q(\rx)\left(\frac{p(\rx|\ry)^\alpha}{p(\rx)^\beta}\right),
    \vspace{-0.7em}
\end{align}
where the proportionality holds because $p(\ry)$ is fixed.

\subsection{Pseudocode for language model guidance}
\label{app:pseudocode_lmg}
\begin{python}
# captioner: Captioning model (returns token log probs)
# lm: Language model (returns token log probs)
# prompt_tokens: Tokenized prompt for language model
# img_embed: Image embedding
# alpha, beta: Cond/uncond exponents from Eq. 4
# max_length: Maximum number of tokens in caption
# BOS: Beginning of sequence token
# EOS: End of sequence token
# NEWLINE: Newline token

tokens = [BOS]
for i in range(0, max_length):
  # Log of Eq. 4.
  lm_log_probs = lm(concat(prompt_tokens, tokens))
  cond_log_probs = captioner(tokens, img_embed)
  uncond_log_probs = captioner(tokens, zeros_like(img_embed))
  scores = lm_log_probs + alpha * cond_log_probs - beta * uncond_log_probs

  # Transfer probability mass from NEWLINE to EOS.
  scores[EOS] = logsumexp([scores[EOS], scores[NEWLINE]])
  scores[NEWLINE] = -inf

  # Greedily take the next token.
  next_token = argmax(scores)
  tokens.append(next_token)
  if next_token == EOS: break
\end{python}

\subsection{Manually written prompts}
\noindent Below, we include the manually written prompts that we use in our language model guidance experiments. Each caption is separated by two newlines.

\label{app:captions}
\subsubsection{Descriptive caption prompt}
\begin{lstlisting}
a bathroom with goldenrod circular patterned tiles contains a toilet bidet sink mirror tissue dispenser and hairdryer\n
donuts being sorted on the conveyor belt of a device labeled donut robot in an industrial kitchen\n
a green glass mug containing 3 toothbrushes and 1 tube of toothpaste sitting on a windowsill\n
a man wearing sunglasses and a gray shirt poses with a woman wearing a white shirt next to a giraffe with a fence behind them\n
a snow covered wooden bench in front of a fence with snow covered evergreen plants behind it\n
two white horses pull a plow with a man in a white shirt and cyan cap and a man in a red shirt with sunglasses behind them next to a fence under a sky with cumulus clouds\n
a man in a blue shirt and a small child in a red striped shirt play frisbee next to trees in a park\n
a black clock tower with a lit up white clock face with roman numerals in front of a dilapidated five story warehouse after dusk\n
a decorative pool flanked by palm trees in front of a stone clock tower next to a large ten story building with a bright advertisement on top in a city at night\n
cows with gray bodies and white heads eating grass on a hill with a foggy mountain in the background\n
\end{lstlisting}
\subsubsection{Counting prompt}
\begin{lstlisting}
a photo of four clouds\n
a photo of one cat\n
a photo of three horses\n
a photo of seven candles\n
a photo of sixteen keys\n
a photo of one rat\n
a photo of five carrot sticks\n
a photo of one turtle\n
a photo of two boats\n
a photo of one orange\n
a photo of nine books\n
a photo of ten fingers\n
a photo of twelve eggs\n
a photo of one microwave\n
a photo of two children\n
a photo of six leaves\n
a photo of two monitors\n
a photo of one toilet\n
a photo of one house\n
a photo of five pairs of pants\n
a photo of eight apples\n
a photo of eleven stars\n
a photo of one hat\n
a photo of two chairs\n
a photo of seven coins\n
a photo of three birds\n
\end{lstlisting}

\subsection{Difference between attention pooling and bottleneck CoCa architecture}
\begin{figure}[h]
    \centering
    \includegraphics{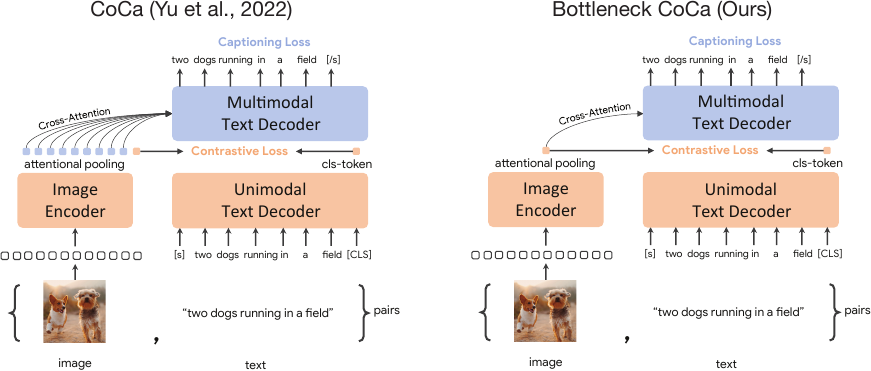}
    \caption{Comparison of CoCa architecture introduced by Yu et al.~\citet{yu2022coca} (left) with our bottleneck CoCa architecture (right).}
    \label{fig:coca_architcture}
\end{figure}

Yu et al.~\citet{yu2022coca} perform attentional pooling over the token representations of the image encoder and pass the resulting tokens into the multimodal text decoder (Figure~\ref{fig:coca_architcture} left). By contrast, our bottleneck architecture uses the same embedding for the contrastive loss and multimodal text decoder (Figure~\ref{fig:coca_architcture} right). We create this bottleneck because a goal of our work is to invert contrastive embeddings, producing a caption that lies close to the contrastive image embedding when it is embedded by the text encoder. As we show below in Appendix~\ref{app:coca_no_bottleneck}, this bottleneck is not necessary for CFG to yield improvements. The attention pooling architecture is equally compatible with our approach and yields slightly better performance.

\label{app:architecture}

\section{Additional experimental results}
\subsection{Attention pooling CoCa architecture}
\label{app:coca_no_bottleneck}

\begin{figure*}[h]
\begin{center}
\includegraphics[width=0.82\linewidth]{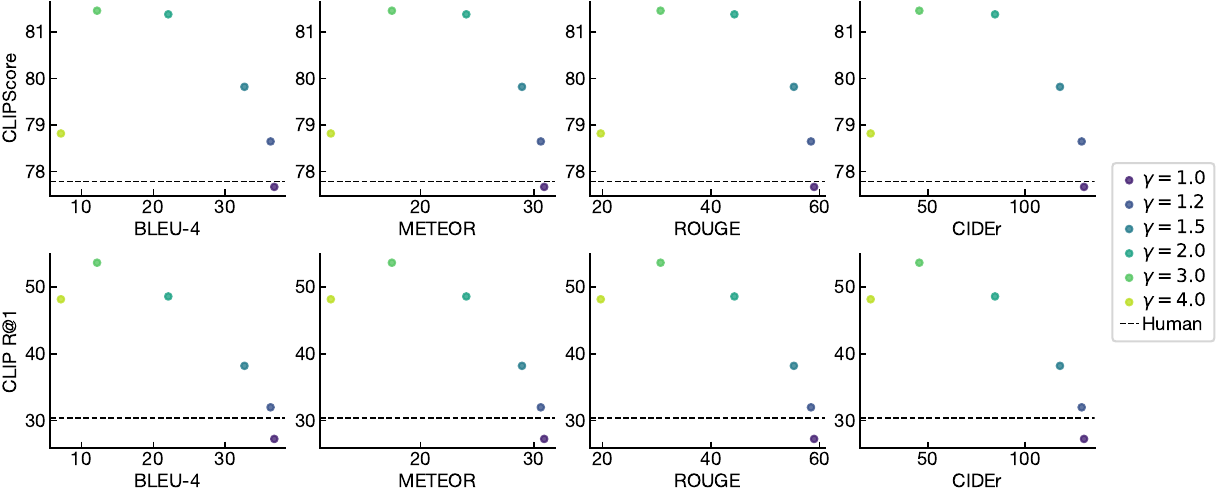}
\end{center}
\vspace{-0.5em}
\caption{Effect of classifier-free guidance on captioning metrics with the attention pooling CoCa model. All points reflect the same fine-tuned model; each color represents a $\gamma$ value used to decode. Models are evaluated with different guidance scales $\gamma$, using reference-free captioning metrics based on CLIP ViT-B/32 (y-axes; top: CLIPScore, bottom: recall@1) and reference-based captioning metrics (x-axes). The dashed line reflects the value of the reference-free captioning metric for the ground-truth captions obtained from MS-COCO. See Figure~\ref{fig:clip_vs_cider} for results with the bottleneck model.}
\label{fig:clip_vs_cider_no_bottleneck}
\vspace{-1em}
\end{figure*}

Classifier-free guidance yields similar qualitative results (and slightly better quantiative results) when using the standard CoCa architecture with attention pooling (Figure~\ref{fig:coca_architcture} left) rather than the bottleneck architecture used in the main text (Figure~\ref{fig:coca_architcture} right). We fine-tune CoCa-Base for 20,000 steps with a max learning rate of $1\times 10^{-5}$ and a conditioning masking proportion of 0.5, following the same procedure that gave the near-optimal bottleneck model described in Section~\ref{sec:models_and_training}. %
Figure~\ref{fig:clip_vs_cider_no_bottleneck} plots reference-based metrics on the x-axis and reference-free metrics on the y-axis, showing a similar trade-off to Figure~\ref{fig:clip_vs_cider}. Table~\ref{tab:clip_vs_cider_no_bottleneck} provides quantitative results demonstrating that the attention pooling architecture performs slightly better across both reference-based and reference-free evaluations. Nonetheless, we adopt the bottleneck architecture for our main experiments for the reasons described in Appendix~\ref{app:architecture} above.

\begin{table*}[h]
\begin{center}
\footnotesize
\setlength{\tabcolsep}{5pt}
\begin{tabular}{|l|rrrrl|lrrr|l|}
\hline
\multicolumn{1}{|c}{} & \multicolumn{5}{|c|}{Reference-Based Metrics} & \multicolumn{4}{|c|}{Reference-Free Metrics} &\\
\multicolumn{1}{|c|}{Model}  &\multicolumn{1}{c}{BLEU-4} &\multicolumn{1}{c}{METEOR} &\multicolumn{1}{c}{ROUGE} & \multicolumn{1}{c}{CIDEr} & \multicolumn{1}{c|}{RefOnlyCLIP-S} & \multicolumn{1}{c}{CLIP-S} &\multicolumn{1}{c}{R@1} &\multicolumn{1}{c}{R@5} &\multicolumn{1}{c|}{R@10}&\multicolumn{1}{c|}{RefCLIP-S}
\\ \hline \hline
          Bottleneck ($\gamma = 1.0$) &    \textbf{36.1} & \textbf{30.5} &     \textbf{58.2} &  \textbf{126.1} &                        \textbf{0.900} &                0.775 &              26.5 &              51.9 &               64.1 &                    0.830 \\
          Bottleneck ($\gamma = 1.2$) &    35.1 & 30.0 &     57.5 &  124.1 &                        0.899 &                0.785 &              31.3 &              57.4 &               69.3 &                    0.835 \\
          Bottleneck ($\gamma = 1.5$) &    31.5 & 28.4 &     54.4 &  113.2 &                        0.891 &                0.796 &              36.6 &              64.0 &               75.0 &                    \textbf{0.838} \\
          Bottleneck ($\gamma = 2.0$) &    20.9 & 23.3 &     43.0 &   78.6 &                        0.862 &                \textbf{0.808} &              44.6 &              71.7 &               81.7 &                    0.831 \\
          Bottleneck ($\gamma = 3.0$) &    11.5 & 17.1 &     29.4 &   41.7 &                        0.820 &                \textbf{0.808} &              \textbf{49.4} &              \textbf{75.7} &               \textbf{84.7} &                    0.811 \\
          Bottleneck ($\gamma = 4.0$) &     6.5 & 12.3 &     18.4 &   17.3 &                        0.766 &                0.782 &              44.7 &              71.3 &               80.9 &                    0.771 \\
          \hline
            Att. Pooling ($\gamma = 1.0$) &    \textbf{36.8} &    \textbf{30.9} &     \textbf{59.0} &  \textbf{130.3} &                        \textbf{0.901} &                0.777 &              27.2 &              52.7 &               64.6 &                    0.832 \\
            Att. Pooling ($\gamma = 1.2$) &    36.3 &    30.6 &     58.4 &  129.1 &                       \textbf{0.901} &                0.786 &              32.0 &              58.0 &               69.4 &                    0.837 \\
            Att. Pooling ($\gamma = 1.5$) &    32.7 &    29.0 &     55.3 &  118.0 &                        0.892 &                0.798 &              38.2 &              64.9 &               75.6 &                    \textbf{0.840} \\
            Att. Pooling ($\gamma = 2.0$) &    22.1 &    24.0 &     44.3 &   84.6 &                        0.861 &                0.814 &              48.6 &              73.7 &               83.5 &                    0.833 \\
            Att. Pooling ($\gamma = 3.0$) &    12.2 &    17.5 &     30.7 &   45.7 &                        0.816 &                \textbf{0.815} &              \textbf{53.6} &              \textbf{78.2} &               \textbf{86.0} &                    0.812 \\
            Att. Pooling ($\gamma = 4.0$) &     7.2 &    12.1 &     19.7 &   20.7 &                        0.767 &                0.788 &              48.2 &              72.1 &               80.1 &                    0.773\\
        \hline
\end{tabular}
\end{center}
\caption{Quantitative comparison of results obtained with bottleneck and attention pooling architectures.}
\label{tab:clip_vs_cider_no_bottleneck}
\vspace{-1em}
\end{table*}
\FloatBarrier

\subsection{Quantitative assessment of specificity}
\label{app:quantitative_specificity}

\subsubsection{Evaluation on Stanford Dogs}

\begin{table}[h]
\begin{center}
\begin{tabular}{|l|r|r|}
\hline
$\gamma$ & \% Containing Breed & \% Breeds Correct\\
\hline\hline
1.0 & 1.9 & 61.7\\
1.2 & 6.2 & 69.0\\
1.5 & 15.9 & 69.7\\
2.0 & 42.4 & 58.5\\
3.0 & 67.0 & 53.3\\
\hline
\end{tabular}
\end{center}
\caption{We generate captions for the 8,580 captions in the Stanford Dogs test set and measure the percentage of the captions that contain the name of one of the 120 dog classes (``\% Containing Breed'') and the percentage of those captions where that name is correct (``\% Breeds Correct'').}
\label{tab:stanford_dogs}
\end{table}
\FloatBarrier

\subsubsection{Human evaluation}
We performed a human evaluation in which we presented crowdsourcing workers with each image and the two possible captions. We experimented with asking subjects to pick the better caption and the more descriptive caption either on different forms or the same form. When asking subjects to pick only the better caption, we provided the following instructions:

\begin{quote}
Please answer a survey about comparing the quality of two captions for each image.

We will present to you an image and ask which caption is better.
\end{quote}

When asking subjects to pick the more descriptive caption, we instead provided the following instructions:

\begin{quote}
Please answer a survey about comparing the descriptiveness of two captions for each image.

We will present to you an image and ask which caption is a more detailed description of the image. Please ignore grammatical errors that do not affect readability.
\end{quote}

When asking both questions simultaneously, we instructed the subjects as follows:

\begin{quote}
Please answer a survey about comparing two captions for each image.

We will present to you an image and ask a couple questions about:

1) descriptiveness: "Which caption is a more detailed description of the image?"

2) quality: "Which caption is better?"
\end{quote}

In each case, subjects saw the image along with the two captions (in random order) as well as the option "I'm indifferent." Subjects clicked the radio button next to their preferred choice. We excluded 55 images for which the captions generated without guidance and at $\gamma = 2.0$ were identical, resulting in a total of 4,945 images. We obtained a single rating for each image in each condition.

Results are shown in Table~\ref{tab:human_eval}. When we asked which caption was ``better'' and which was ``more descriptive'' in separate surveys, we found that subjects preferred each caption at a statistically indistinguishable rate. When we asked subjects to pick the ``better'' and ``more descriptive'' captions in the same survey, we found that $\gamma = 1.0$ was more likely to be chosen as ``better'' whereas $\gamma = 2.0$ was more likely to be chosen as ``more specific.'' Comparing the odds ratios obtained with the two ways of posing the questions using Fisher's exact test, we find that the difference between them is statistically significant (``better'': $p = 0.004$; ``more descriptive'': $p = 0.01$) indicating that human judgments are significantly affected by whether the questions are posed on the same form or separately.

\begin{table}[h]
\begin{center}
\begin{tabular}{|lrrrl|}
\hline
Question & $\gamma = 1.0$ & $\gamma = 2.0$ & Indifferent & $p$-value\\
\hline\hline
\multicolumn{5}{|l|}{\textit{Separate forms:}}\\
Better & \textbf{48.0\%} (2375) & \textbf{49.8\%} (2461) & 2.2\% (109) & $p = 0.22$\\
More descriptive & \textbf{47.7\%} (2359) & \textbf{49.5\%} (2446) & 2.8\% (140) & $p = 0.21$\\
\hline
\multicolumn{5}{|l|}{\textit{Same form:}}\\
Better & \textbf{50.5\%} (2497) & 46.6\% (2306) & 2.9\% (142) & $p = 0.006$\\
More descriptive & 45.8\% (2265) & \textbf{52.7\%} (2606) & 1.5\% (74) & $p = 10^{-6}$\\
\hline
\end{tabular}
\end{center}
\caption{Human evaluation results. We report the percentage and overall number of the 5,000 MS-COCO Karpathy test set images where subjects preferred captions generated at $\gamma = 1.0$ or $\gamma = 2.0$ or were indifferent, as well as the $p$-value for the null hypothesis that users are equally likely to select the captions generated at $\gamma = 1.0$ and $\gamma = 2.0$, computed by a binomial test. When $p < 0.05$, we bold-face the best result in each row. Otherwise, we bold-face both results.}
\label{tab:human_eval}
\end{table}
\FloatBarrier

\subsection{Reference-free metrics with retrieval models}
\label{app:more_retrieval_models}

In Table~\ref{tab:coca_model_metrics}, we show cosine similarity between generated captions and image embeddings and caption$\to$image retrieval accuracy for the CoCa 2B model and the CoCa-Base model fine-tuned on MS-COCO that was used to generate the captions. In both cases, we find that $\gamma > 1$ yields much better metrics than no guidance. Retrieval accuracies (but not cosine similarities) are directly comparable across models;  both models offer better retrieval accuracy than CLIP ViT-B/32.

\begin{table}[h]
\begin{center}
\begin{tabular}{|l|lrrr|lrrr|}
\hline
\multicolumn{1}{|c}{} & \multicolumn{4}{|c|}{CoCa 2B} & \multicolumn{4}{|c|}{Captioning Model (CoCa Base)}\\
$\gamma$ & Cos. & R@1 & R@5 & R@10 & Cos. & R@1 & R@5 & R@10 \\
\hline\hline
            1.0 &                    0.125 &         40.1 &         65.3 &          75.1 &                             0.843 &                  49.4 &                  75.0 &                   84.1 \\
            1.2 &                    0.128 &         46.5 &         72.0 &          80.3 &                             0.859 &                  56.2 &                  80.1 &                   88.1 \\
            1.5 &                    0.131 &         55.5 &         78.9 &          86.4 &                             0.877 &                  64.6 &                  85.9 &                   91.5 \\
            2.0 &                    \textbf{0.135} &         64.9 &         86.4 &          91.3 &                             0.887 &                  73.0 &                  91.6 &                   95.3 \\
            3.0 &                    0.134 &         \textbf{66.5} &         \textbf{87.0} &          \textbf{91.4} &                             \textbf{0.890} &                  \textbf{77.7} &                  \textbf{92.4} &                   \textbf{95.8} \\
            4.0 &                    0.126 &         60.3 &         81.8 &          87.5 &                             0.875 &                  74.7 &                  90.1 &                   94.0 \\
\hline
\end{tabular}
\end{center}
\caption{CFG improves caption$\to$image retrieval in the embedding spaces of CoCa models on MS-COCO. ``Cos.'' = mean cosine similarity between the image and text embeddings.}
\label{tab:coca_model_metrics}
\end{table}

\clearpage
\section{Additional examples}
\label{app:examples}

\begin{figure*}[h!]
\vspace{-1em}
\begin{center}
\tablefont
\setlength{\tabcolsep}{0pt}
\RaggedRight
\begin{tabular}{p{0.142\linewidth}p{0.142\linewidth}p{0.142\linewidth}p{0.142\linewidth}p{0.142\linewidth}p{0.142\linewidth}p{0.142\linewidth}}
\hfil\includegraphics[width=0.8\linewidth]{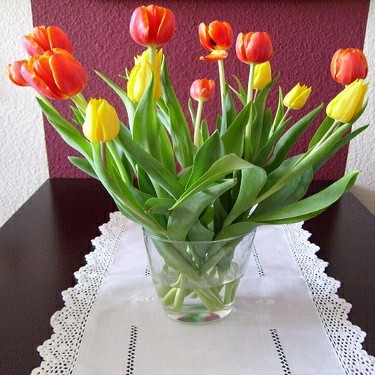} &
\hfil\includegraphics[width=0.8\linewidth]{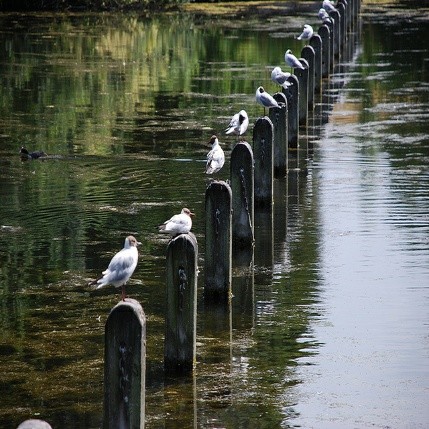} &
\hfil\includegraphics[width=0.8\linewidth]{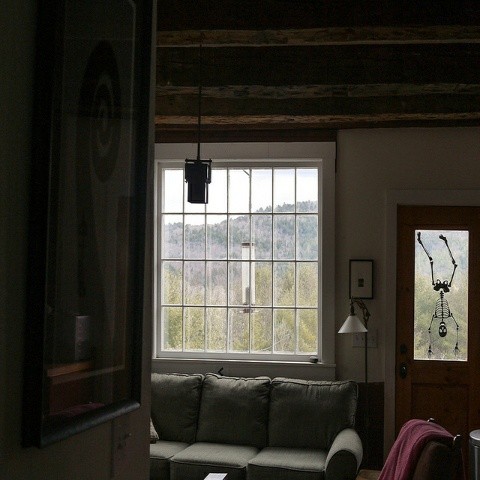} &
\hfil\includegraphics[width=0.8\linewidth]{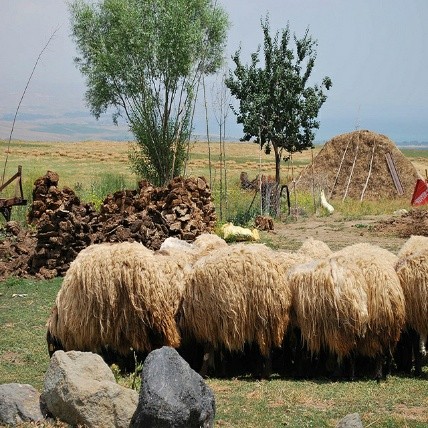} &
\hfil\includegraphics[width=0.8\linewidth]{figures/example_images_appendix/4.jpg} &
\hfil\includegraphics[width=0.8\linewidth]{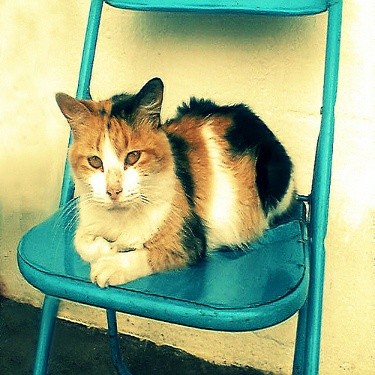} &
\hfil\includegraphics[width=0.8\linewidth]{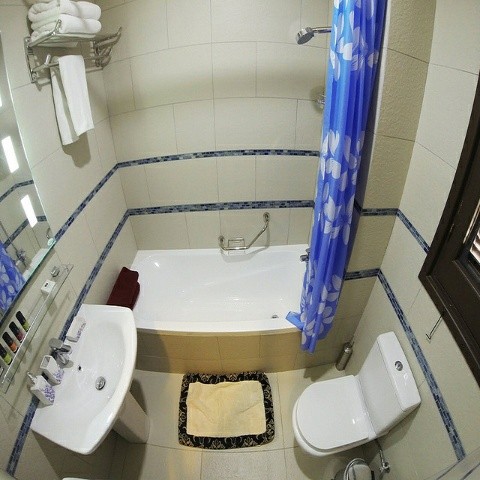}
\\
{\setlength{\tabcolsep}{1pt}\begin{tabular}[t]{rp{0.74\linewidth}}
$\gamma$=1.0& a vase filled with red and yellow flowers\\
$\gamma$=1.5& tulips in a clear vase on a table\\
$\gamma$=2.0& tulips in a clear glass vase on a tablecloth\\
$\gamma$=3.0& tulips in a clear punchov glass setting on doily\\
GT& Fresh red and yellow tulips in a vase.
\end{tabular}} &
{\setlength{\tabcolsep}{1pt}\begin{tabular}[t]{rp{0.74\linewidth}}
$\gamma$=1.0& a group of birds standing on top of a wooden post\\
$\gamma$=1.5& seagulls lined up on posts in a lake\\
$\gamma$=2.0& seagulls lined up along a pond line\\
$\gamma$=3.0& seagulls lined up along posts in shallow water\\
GT& Wood post lined up in the water with birds perched on them. 
\end{tabular}} &
{\setlength{\tabcolsep}{1pt}\begin{tabular}[t]{rp{0.74\linewidth}}
$\gamma$=1.0& a living room with a couch and a table\\
$\gamma$=1.5& a living room with a couch and a window\\
$\gamma$=2.0& living room with large window overlooking woods\\
$\gamma$=3.0& livingroom with view out the window\\
GT& A living room in a remotely located home.
\end{tabular}} &
{\setlength{\tabcolsep}{1pt}\begin{tabular}[t]{rp{0.74\linewidth}}
$\gamma$=1.0& a herd of sheep grazing in a field\\
$\gamma$=1.5& a herd of sheep grazing in a field\\
$\gamma$=2.0& sheep are gathered in a field near piles of hay\\
$\gamma$=3.0& bales of sheep are gathered in formation near rocks\\
GT& A herd of sheep standing on top of a grass covered field.
\end{tabular}} &
{\setlength{\tabcolsep}{1pt}\begin{tabular}[t]{rp{0.74\linewidth}}
$\gamma$=1.0& a pair of skis sitting on a tiled floor\\
$\gamma$=1.5& pair of skis and ski boots on tiled floor\\
$\gamma$=2.0& skis and pair of skis on linoleum floor\\
$\gamma$=3.0& skis and pair of bottle opener sit on vct floor\\
GT& Skis and ski boots sit together on a tiled floor.
\end{tabular}} &
{\setlength{\tabcolsep}{1pt}\begin{tabular}[t]{rp{0.74\linewidth}}
$\gamma$=1.0& a cat sitting on a blue chair with a white wall behind it\\
$\gamma$=1.5& a cat sitting on a blue chair outside\\
$\gamma$=2.0& calico colored cat sitting on blue chair outside\\
$\gamma$=3.0& calico colored cat sitting on blue metal chair\\
GT& A furry cat sits on a blue chair. 
\end{tabular}} &
{\setlength{\tabcolsep}{1pt}\begin{tabular}[t]{rp{0.74\linewidth}}
$\gamma$=1.0& a bathroom with a toilet sink and bathtub\\
$\gamma$=1.5& a bathroom with blue and white tiles and a blue and white towel\\
$\gamma$=2.0& bathroom with blue accents and blue and white towels\\
$\gamma$=3.0& spotless uncroom bathroom with blue accents fisheye fisheye fisheye fisheye fisheyemmangles viewersquallly fisheye and fisheye lens\\
GT& Bathroom with white pedestal sink, bathtub and shower, and commode.
\end{tabular}}\\
\\
\hfil\includegraphics[width=0.8\linewidth]{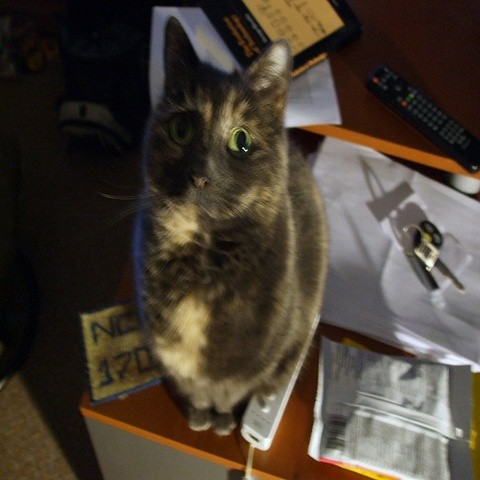} &
\hfil\includegraphics[width=0.8\linewidth]{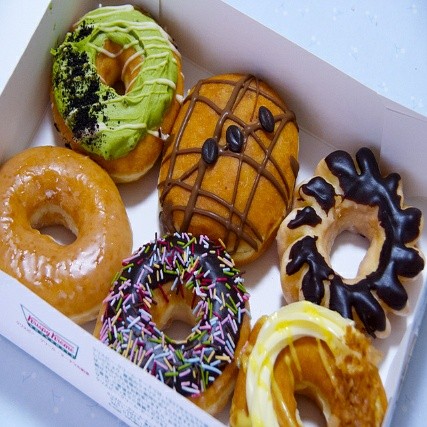} &
\hfil\includegraphics[width=0.8\linewidth]{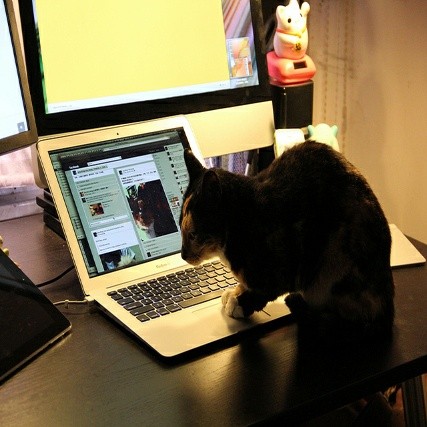} &
\hfil\includegraphics[width=0.8\linewidth]{figures/example_images_appendix/10.jpg} &
\hfil\includegraphics[width=0.8\linewidth]{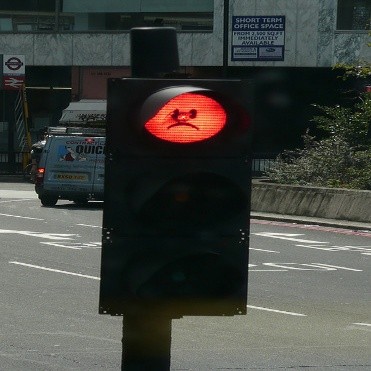} &
\hfil\includegraphics[width=0.8\linewidth]{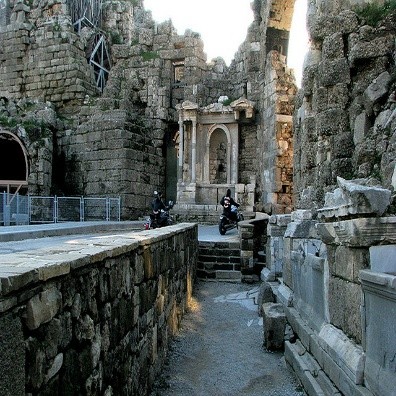} &
\hfil\includegraphics[width=0.8\linewidth]{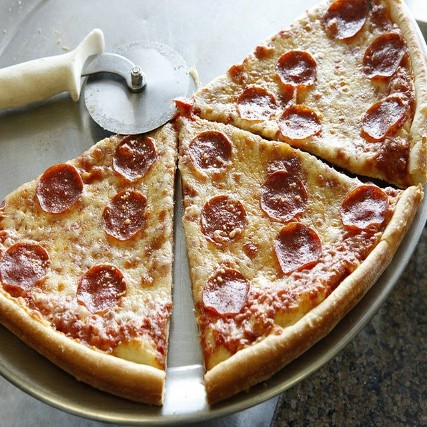}
\\
{\setlength{\tabcolsep}{1pt}\begin{tabular}[t]{rp{0.74\linewidth}}
$\gamma$=1.0& a cat sitting on top of a desk next to a box\\
$\gamma$=1.5& a cat sitting on top of a desk\\
$\gamma$=2.0& a cat sitting on top of files on a cabinet\\
$\gamma$=3.0& tortoiseshell mittedtabkat sitting inquisitive on papers\\
GT& Cat sitting next to remote control on small counter.
\end{tabular}} &
{\setlength{\tabcolsep}{1pt}\begin{tabular}[t]{rp{0.74\linewidth}}
$\gamma$=1.0& a box of assorted donuts with a variety of toppings\\
$\gamma$=1.5& a box of assorted donuts with different toppings\\
$\gamma$=2.0& six glazed and chocolate sprinkled doughnuts in a box\\
$\gamma$=3.0& krispy box of dozen glazed and decorated doughnuts\\
GT& Half a dozen donuts from Krispy Kreme of various different flavors.
\end{tabular}} &
{\setlength{\tabcolsep}{1pt}\begin{tabular}[t]{rp{0.74\linewidth}}
$\gamma$=1.0& a cat sitting on a desk next to a laptop\\
$\gamma$=1.5& a cat sitting on a desk next to a laptop\\
$\gamma$=2.0& cat sitting on desk looking at lap top screen\\
$\gamma$=3.0& calico laptop sitting on computer desk with calico cat sitting on top of screen\\
GT& A cat standing on top of a laptop computer.
\end{tabular}} &
{\setlength{\tabcolsep}{1pt}\begin{tabular}[t]{rp{0.74\linewidth}}
$\gamma$=1.0& a large brown and black insect on top of a laptop\\
$\gamma$=1.5& a bug sitting on the edge of a laptop\\
$\gamma$=2.0& dragonfly perched on television outside on patio\\
$\gamma$=3.0& dragonfly perched on television outside on cantilever table\\
GT& A bug sitting on the side of a laptop computer.
\end{tabular}} &
{\setlength{\tabcolsep}{1pt}\begin{tabular}[t]{rp{0.74\linewidth}}
$\gamma$=1.0& a red traffic light sitting on the side of a road\\
$\gamma$=1.5& a traffic light with a red pedestrian crossing sign on it\\
$\gamma$=2.0& red traffic light sitting on the side of a street\\
$\gamma$=3.0& pedestrian signal red on a black light pole\\
GT& A red traffic light with a sad face drawn over it.
\end{tabular}} &
{\setlength{\tabcolsep}{1pt}\begin{tabular}[t]{rp{0.74\linewidth}}
$\gamma$=1.0& a stone wall with a clock tower and a stone wall\\
$\gamma$=1.5& ruins of a building with people walking around\\
$\gamma$=2.0& ruins at a castle in turkey\\
$\gamma$=3.0& ruins at diocletianopolis roman ruins\\
GT& A city made out of stone brick with large arches.
\end{tabular}} &
{\setlength{\tabcolsep}{1pt}\begin{tabular}[t]{rp{0.74\linewidth}}
$\gamma$=1.0& a pizza that is sitting on a pan\\
$\gamma$=1.5& pepperoni pizza on metal pan with cutter\\
$\gamma$=2.0& pepperoni pizza on metal pan with cutter\\
$\gamma$=3.0& pepperoni steel traybake pepperoni steel tray pizza cutter pepperoni steel tray\\
GT& A pan with three pieces of pepperoni pizza.
\end{tabular}}
\\
\\
\hfil\includegraphics[width=0.8\linewidth]{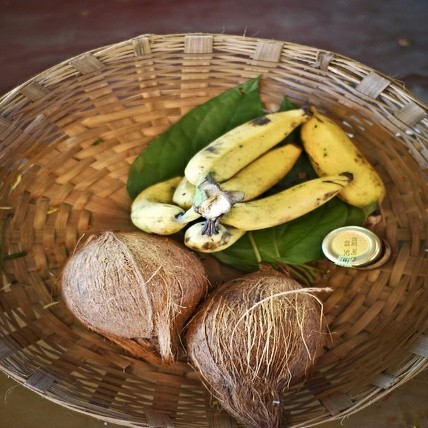} &
\hfil\includegraphics[width=0.8\linewidth]{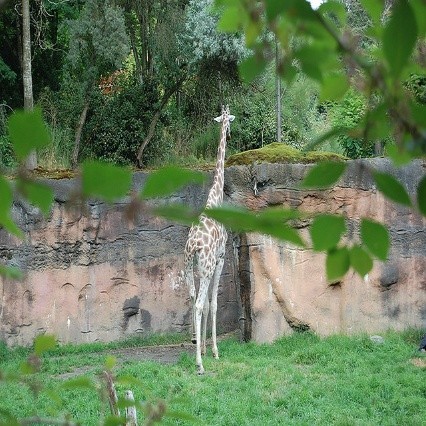} &
\hfil\includegraphics[width=0.8\linewidth]{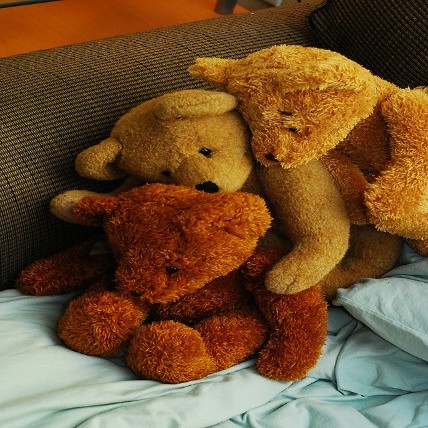} &
\hfil\includegraphics[width=0.8\linewidth]{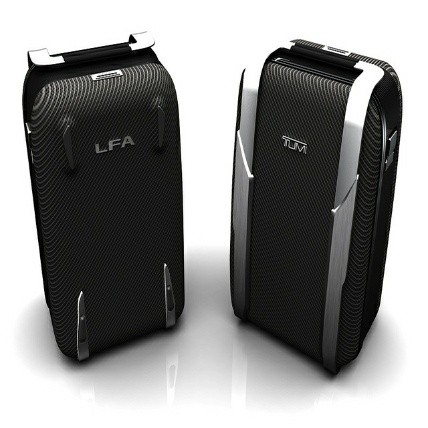} &
\hfil\includegraphics[width=0.8\linewidth]{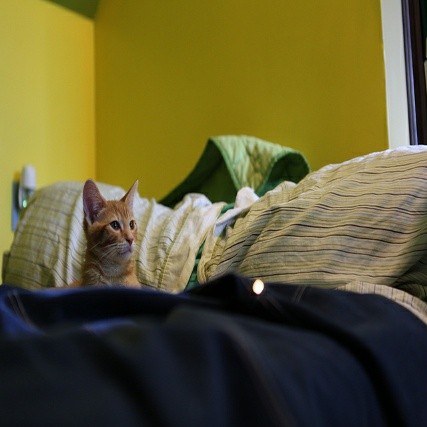} &
\hfil\includegraphics[width=0.8\linewidth]{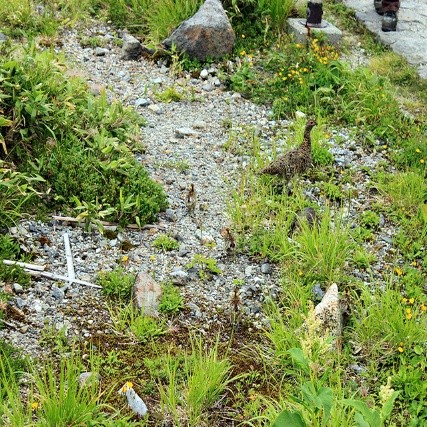} &
\hfil\includegraphics[width=0.8\linewidth]{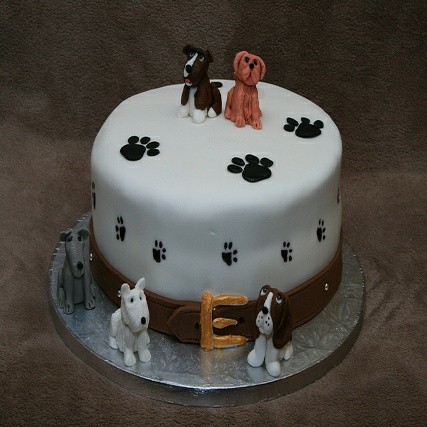}
\\
{\setlength{\tabcolsep}{1pt}\begin{tabular}[t]{rp{0.74\linewidth}}
$\gamma$=1.0& a basket of bananas and coconuts on a table\\
$\gamma$=1.5& coconut and bananas in a basket with a banana inside\\
$\gamma$=2.0& coconut basket with bananas and nuts in it\\
$\gamma$=3.0& coconut basket bananas coconut husknus and husk laid out\\
GT& a basket with a few things of fruit in it
\end{tabular}} &
{\setlength{\tabcolsep}{1pt}\begin{tabular}[t]{rp{0.74\linewidth}}
$\gamma$=1.0& a giraffe standing in a grassy area next to a rock wall\\
$\gamma$=1.5& a giraffe standing in a grassy area next to a rock wall\\
$\gamma$=2.0& giraffe standing in enclosure near trees and rock wall\\
$\gamma$=3.0& girafe confined motionless zoo confined wild confined into captivity\\
GT& A giraffe walking through a lush green field.
\end{tabular}} &
{\setlength{\tabcolsep}{1pt}\begin{tabular}[t]{rp{0.74\linewidth}}
$\gamma$=1.0& a group of teddy bears sitting on a bed\\
$\gamma$=1.5& three teddy bears sitting on a bed together\\
$\gamma$=2.0& four teddy bears sitting on a bed together\\
$\gamma$=3.0& cuddling teddy bears lay piled on a sofa\\
GT& Three different teddy bear on a blanket on a chair.
\end{tabular}} &
{\setlength{\tabcolsep}{1pt}\begin{tabular}[t]{rp{0.74\linewidth}}
$\gamma$=1.0& two black suitcases are sitting next to each other\\
$\gamma$=1.5& two suitcases with wheels on white background\\
$\gamma$=2.0& two suitcases facing each other 3d illustration\\
$\gamma$=3.0& cgi suitcases rendered cgi cgi cgi looks like luggages cgi cgie cgih cgih cgih cgih cgih cgih cgih cgih cgih cgih cgih cgih cgih travelshpinky like nexushxm gif 3dding\\
GT& A couple of pieces of very nice looking luggage.
\end{tabular}} &
{\setlength{\tabcolsep}{1pt}\begin{tabular}[t]{rp{0.74\linewidth}}
$\gamma$=1.0& a cat sitting on a bed next to a blanket\\
$\gamma$=1.5& a cat sitting on a bed under a blanket\\
$\gamma$=2.0& a tabby kitten sitting on top of a comforter on a bed\\
$\gamma$=3.0& tabby kitten sitting on uncovered rumple drapes on unmade unmade unmade bed\\
GT& A brown and white cat lying on a bed
\end{tabular}} &
{\setlength{\tabcolsep}{1pt}\begin{tabular}[t]{rp{0.74\linewidth}}
$\gamma$=1.0& a bird is standing on the ground in the grass\\
$\gamma$=1.5& weeds and rocks in a grassy area with dirt\\
$\gamma$=2.0& weeds and rocks litter a gravel path in a grassy area\\
$\gamma$=3.0& weeds and gravel strewn away along gravel trail strewn with bird rocks\\
GT& A bird walking through some gravel as its baby chicks follow.
\end{tabular}} &
{\setlength{\tabcolsep}{1pt}\begin{tabular}[t]{rp{0.74\linewidth}}
$\gamma$=1.0& a cake with a dog and horse on it\\
$\gamma$=1.5& a cake with dogs and horses on it\\
$\gamma$=2.0& cake decorated dog horse and dog motif with three horses\\
$\gamma$=3.0& cake puppy horse dog dog and cats decorated for a first birthday\\
GT& A cake that has paw prints and miniatures dogs on it.
\end{tabular}}
\\
\\
\hfil\includegraphics[width=0.8\linewidth]{figures/example_images_appendix/21.jpg} &
\hfil\includegraphics[width=0.8\linewidth]{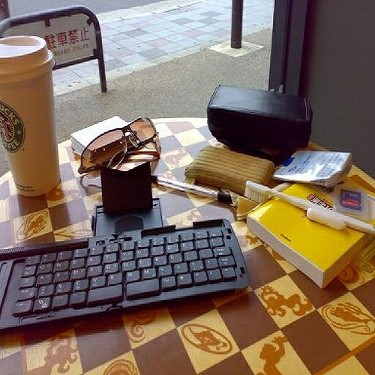} &
\hfil\includegraphics[width=0.8\linewidth]{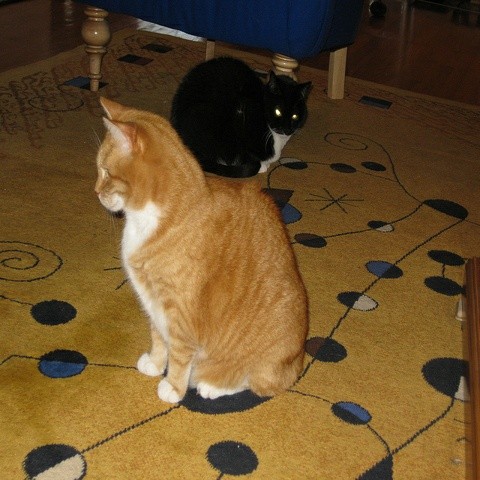} &
\hfil\includegraphics[width=0.8\linewidth]{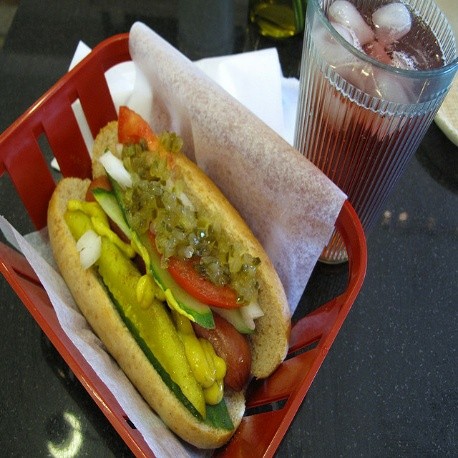} &
\hfil\includegraphics[width=0.8\linewidth]{figures/example_images_appendix/25.jpg} &
\hfil\includegraphics[width=0.8\linewidth]{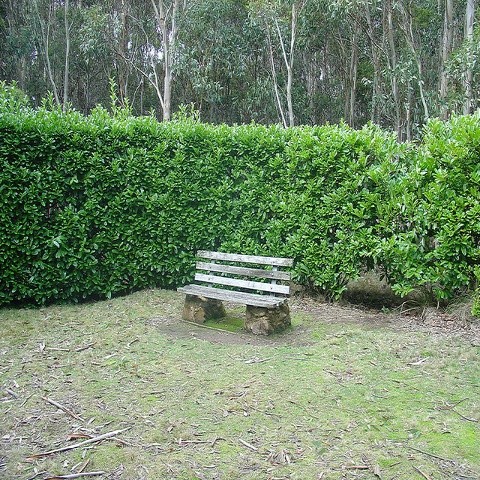} &
\hfil\includegraphics[width=0.8\linewidth]{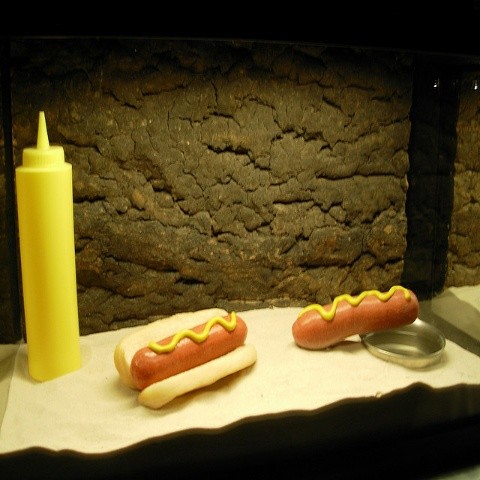}
\\
{\setlength{\tabcolsep}{1pt}\begin{tabular}[t]{rp{0.74\linewidth}}
$\gamma$=1.0& a city street with a clock tower and cars\\
$\gamma$=1.5& a city street at night with cars and buildings\\
$\gamma$=2.0& cars are driving down a busy city street at night\\
$\gamma$=3.0& ginza at night with cars lights and edifice in asia\\
GT& The traffic and people on a commercial street corner at night
\end{tabular}} &
{\setlength{\tabcolsep}{1pt}\begin{tabular}[t]{rp{0.74\linewidth}}
$\gamma$=1.0& a table with a keyboard a cup of coffee and a keyboard\\
$\gamma$=1.5& a keyboard coffee cup and glasses on a table\\
$\gamma$=2.0& keyboard coffee sunglasses pen and cup on outdoor table\\
$\gamma$=3.0& keyboard coffee sunglasses pen wallet keyboard starbucks cup on outdoor table\\
GT& a keyboard on a table with a toothbrush a book some sunglasses and coffee
\end{tabular}} &
{\setlength{\tabcolsep}{1pt}\begin{tabular}[t]{rp{0.74\linewidth}}
$\gamma$=1.0& two cats sitting on a rug in a room\\
$\gamma$=1.5& two cats sitting on a rug in a room\\
$\gamma$=2.0& two cats sitting on rug in room with orange carpet\\
$\gamma$=3.0& cats sitting next to each other on patterned carpet\\
GT& A black cat and an orange cat are sitting on the floor.
\end{tabular}} &
{\setlength{\tabcolsep}{1pt}\begin{tabular}[t]{rp{0.74\linewidth}}
$\gamma$=1.0& a sandwich and a drink in a basket on a table\\
$\gamma$=1.5& a sandwich and a drink in a basket on a table\\
$\gamma$=2.0& sandwich basket with drink and pickle relish\\
$\gamma$=3.0& sandwich basket drink relish relish pickle hot dog and drink\\
GT& A hotdog with toppings served in a red basket
\end{tabular}} &
{\setlength{\tabcolsep}{1pt}\begin{tabular}[t]{rp{0.74\linewidth}}
$\gamma$=1.0& a plate of food with a sandwich and a drink\\
$\gamma$=1.5& tater tots and a sandwich and tater tots are on a paper plate\\
$\gamma$=2.0& tater tots toast and a beer on a restaurant table\\
$\gamma$=3.0& tater tots toast club sandwich tater tots and beer on a restaurant table\\
GT& A tray with a cheese and meat sandwich with tater tots.
\end{tabular}} &
{\setlength{\tabcolsep}{1pt}\begin{tabular}[t]{rp{0.74\linewidth}}
$\gamma$=1.0& a wooden bench sitting in the middle of a forest\\
$\gamma$=1.5& a bench sitting in the middle of a hedge\\
$\gamma$=2.0& hedges and bench in a forested area\\
$\gamma$=3.0& hedges hedge bench hedges bush hedges\\
GT& A bench out by a hedge by the woods 
\end{tabular}} &
{\setlength{\tabcolsep}{1pt}\begin{tabular}[t]{rp{0.74\linewidth}}
$\gamma$=1.0& a hot dog and a mustard bottle on a table\\
$\gamma$=1.5& a hotdog and mustard are on wax paper next to a counter\\
$\gamma$=2.0& hot dog and mustard candles on wax paper\\
$\gamma$=3.0& dug hot dog and mustard candles on wax paper under counter\\
GT& Two hot dogs sitting on top of tissue paper.
\end{tabular}}
\end{tabular}
\end{center}
\vspace{-0.5em}
\caption{Additional examples of captions generated with classifier-free guidance at different strengths.}
\vspace{-0.5em}
\end{figure*}

\end{document}